\documentclass{article}

\PassOptionsToPackage{numbers}{natbib}
\usepackage[final]{neurips_2021}

\usepackage[utf8]{inputenc} % allow utf-8 input
\usepackage[T1]{fontenc}    % use 8-bit T1 fonts
\usepackage{hyperref}       % hyperlinks
\usepackage{url}            % simple URL typesetting
\usepackage{booktabs}       % professional-quality tables
\usepackage{amsfonts}       % blackboard math symbols
\usepackage{nicefrac}       % compact symbols for 1/2, etc.
\usepackage{microtype}      % microtypography
\usepackage{subcaption}
\usepackage{graphicx}
\usepackage{subcaption}
\usepackage{amsmath}
\usepackage{comment}
\usepackage{lipsum}
\usepackage{bbding}
\usepackage{bbding}
\usepackage{pifont}
\usepackage{wasysym}
\usepackage{amssymb}
\usepackage{multirow}
\usepackage[dvipsnames]{xcolor}

\usepackage{import}
\usepackage{xifthen}
\usepackage{pdfpages}
\usepackage{transparent}

\usepackage{enumitem}

\usepackage{xcolor}
\definecolor{dark-blue}{rgb}{0.15,0.15,0.4}
\definecolor{medium-blue}{rgb}{0,0,0.5}
\hypersetup{
  colorlinks, linkcolor={dark-blue},
  citecolor={dark-blue}, urlcolor={medium-blue}
}

\title{Residual Pathway Priors for Soft Equivariance Constraints}

\author{Marc Finzi\footnotemark[1]\\ New York University \And Gregory Benton\footnotemark[1]\\ New York University \And Andrew Gordon Wilson\\
New York University}

\begin{document}

\maketitle

\begin{abstract}
There is often a trade-off between building deep learning systems that are expressive enough to capture the nuances of the reality, and having the right inductive biases for efficient learning. We introduce Residual Pathway Priors (RPPs) as a method for converting hard architectural constraints into soft priors, guiding models towards structured solutions, while retaining the ability to capture additional complexity. Using RPPs, we 
construct neural network priors with inductive biases for equivariances, but without limiting flexibility. We show that RPPs are resilient to approximate or misspecified symmetries, and are as effective as fully constrained models even when symmetries are exact. We showcase the broad applicability of RPPs with dynamical systems, tabular data, and reinforcement learning. In Mujoco locomotion tasks, where contact forces and 
directional rewards violate strict equivariance assumptions, the RPP outperforms baseline model-free RL agents, and also improves the learned transition models for model-based RL.

\end{abstract}

\section{Introduction}
\let\svthefootnote\thefootnote
\let\thefootnote\relax\footnote{$^*$Equal Contribution}
\addtocounter{footnote}{-1}
\let\thefootnote\svthefootnote
\vspace{-6mm}

Central to the expanding application of deep learning to structured data like images, text, audio, sets, graphs, point clouds, and dynamical systems, has been a search for finding the appropriate set of inductive biases to match the model to the data. These inductive biases, such as recurrence \citep{rumelhart1985learning}, local connectivity \citep{lecun1989backpropagation}, equivariance \citep{cohen2016group}, or differential equations \citep{chen2018neural}, reduce the set of explored hypotheses and improve generalization. Equivariance in particular has had a large impact as it allows ruling out a large class of meaningless shortcut features in many distinct domains, such as the ordering of the nodes in graphs and sets or the coordinate system chosen for an image.

A disadvantage of hard coding these restrictions is that this prior knowledge may not match reality. 
A scene may have long range non-local interactions, rotation equivariance may be violated by a preferred camera angle, or a dynamical system may occasionally have discontinuous transitions. In particular, symmetries are delicate. A small perturbation like adding wind breaks the rotational symmetry of a pendulum, and bumpy or tilted terrain could break the translation symmetry for locomotion. In these cases we would like to incorporate our prior knowledge in a way that admits our own ignorance, and allows for the possibility that the world is more complex than we imagined.
We aim to develop an approach that is more general, and can be applied when symmetries are exact, approximate, or non-existent.

The Bayesian framework provides a mechanism for expressing such knowledge through priors. 
In much of the past work on Bayesian neural networks, the relationship between the prior distribution and the functions preferred by the prior is not transparent. While it is easy to specify different variances for different channels, or to use heavy tailed distributions, it is hard know how high level properties meaningfully translate into these low level attributes. 
Ultimately priors should represent our prior \emph{beliefs}, and the beliefs we have are about high level concepts like the locality, independence, and symmetries of the data.

To address the need for more interpretable priors 
we introduce \emph{Residual Pathway Priors} (RPPs), a method for converting hard architectural constraints into soft priors. Practically, RPPs allow us to tackle problems in which perfect symmetry has been violated, but approximate symmetry is still present, as is the case for most real world physical systems.  RPPs have a prior bias towards equivariant solutions, but are not constrained to them.

\begin{figure}
    \centering
    \begin{subfigure}[t]{0.57\textwidth}\centering
        \centering
        \includegraphics[trim={.5cm 0 0 3cm},clip,width=\columnwidth]{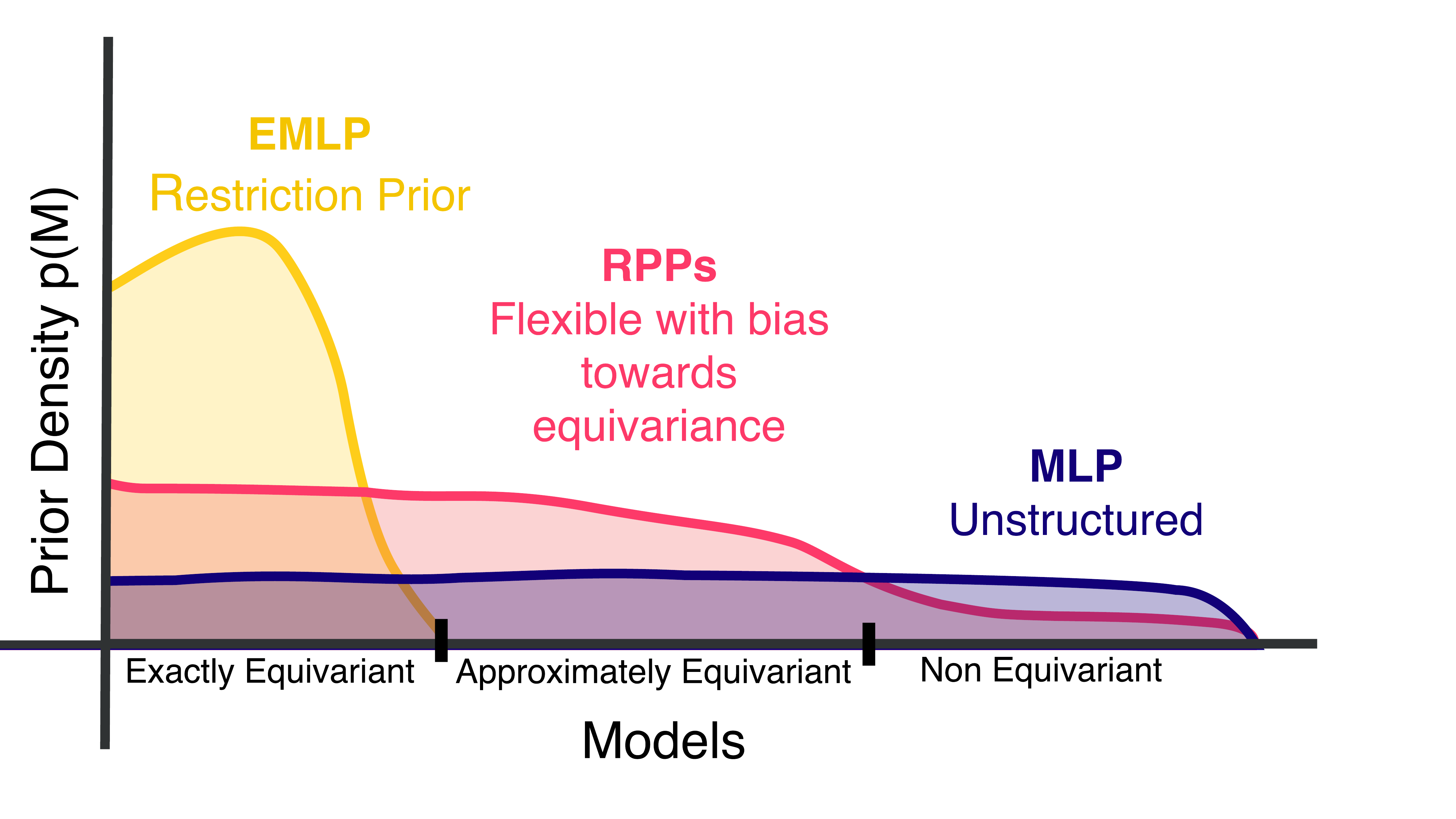}
        \caption{Priors over Equivariant Solutions}
        \label{fig:occam}
    \end{subfigure}%
    \hspace{0.3cm}
    \begin{subfigure}[t]{0.36\textwidth}\centering
        \centering
        \includegraphics[width=\columnwidth]{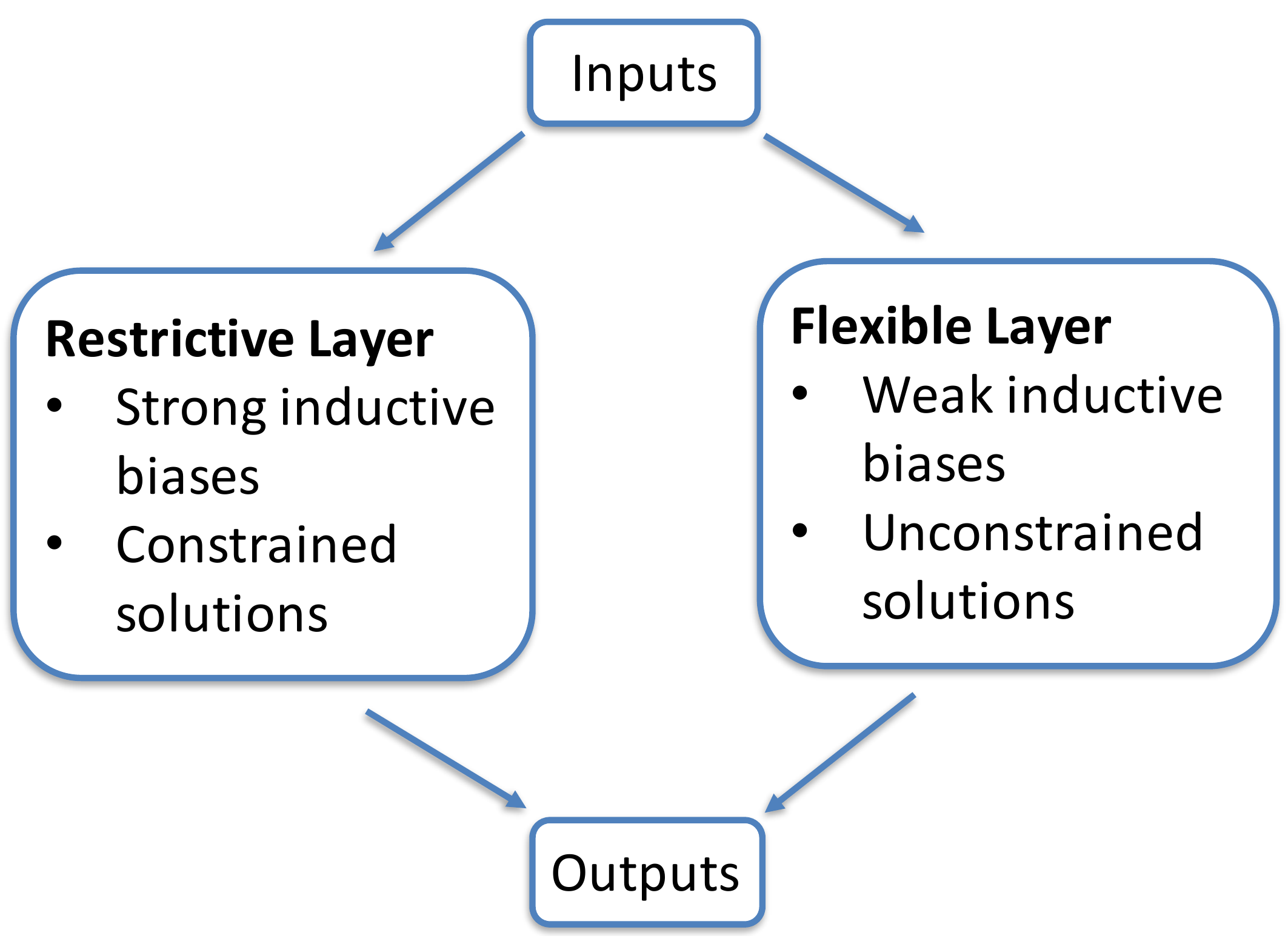}
        \caption{Structure of RPP Models}
        \label{fig:rpp-flow}
    \end{subfigure}
    \caption{
    \textbf{Left:} RPPs encode an Occam's razor approach to modeling. 
    Highly flexible models like MLPs lack the inductive biases to assign high prior mass to relevant solutions for a given problem, while models with strict constraints are not flexible enough to support solutions with only approximate symmetry. For a given problem, we want to use the most constrained model that is consistent with our observations.
    \textbf{Right:} The structure of RPPs. 
    Expanding the layers into a sum of the constrained and unconstrained solutions, while setting the prior to favor the constrained solution, leads to the more flexible layer explaining only the \emph{residual} of what is already explained by the constrained layer.
    }
    \label{fig:probability}
\end{figure}

We use the schematic in Figure \ref{fig:occam} as an approach to model construction \citep{wilson2020bayesian, mackay2003information}. The flexibility of our model is described by what solutions have non-zero prior probability density. The \emph{inductive biases} are described by the distribution of support over solutions. We wish to construct models with inductive biases that assign significant prior mass for solutions we believe to be a priori likely, but without ruling out other solutions we believe to be possible. For example, models constrained to exact symmetries could not fully represent many problems, such as the motion of a pendulum in the presence of wind. Flexible models with poor inductive biases, spread thinly across possible solutions, could express an approximate symmetry, but such solutions are unlikely to be found because of the low prior density. In this sense, we wish to embrace a notion of Occam’s razor such that ``everything should be made as simple as possible, but no simpler''.

As we find with problems in which symmetries exist, highly flexible models with weak inductive biases like MLPs fail to concentrate prior mass around solutions that exhibit any symmetry. 
On the other hand when symmetries are only approximate, the strong restriction biases of constrained models like Equivariant Multi-Layer Perceptrons (EMLP) \citep{finzi2021practical} fail to provide support for the observations. As a middle ground between these two extremes, RPPs combine the inductive biases of constrained models with the flexibility of MLPs to define a model class which excels when data show approximate symmetries, as shown in Figure \ref{fig:rpp-flow}.

In the following sections we introduce our method and show results across a variety of domains. We list our contributions and the accompanying sections below:
\begin{enumerate}[itemsep=0mm]
    \item We propose \emph{Residual Pathway Priors} as a mechanism to imbue models with soft inductive biases, without constraining flexibility.
    \item While our approach is general, we use RPPs to show how to turn hard architectural constraints into soft equivariance priors (Section \ref{sec: methods}).
    \item We demonstrate that RPPs are robust to varying degrees of symmetry (Section \ref{sec: howandwhy}). RPPs perform well under exact, approximate, or misspecified symmetries.
    \item Using RPP on the approximate symmetries in the complex state spaces of the Mujoco locomotion tasks, we improve the performance of model free RL agents (Section \ref{sec: rl-priors}).
\end{enumerate}

We provide a PyTorch implementation of residual pathway priors at \\ \url{https://github.com/mfinzi/residual-pathway-priors}.

\section{Related Work}
The challenge of equivariant models not being able to fully fit the data has been identified in a number of different contexts, and with different application specific adjustments to mitigate the problem. 
\citet{liu2018intriguing} observe that convolutional networks can be extremely poor at tasks that require identifying or outputting spatial locations in an image as a result of the translation symmetry. The authors solve the problem by concatenating a coordinate grid to the input of the convolution layer. Constructing translation and rotation equivariant GCNNs, \citet{weiler2019general} find that in order to get the best performance on CIFAR-10 and STL-10 datasets which have a preferred camera orientation, they must break the symmetry, which they do by using equivariance to progressively smaller subgroups in the later layers. \citet{bogatskiy2020lorentz} go to great lengths to construct Lorentz group equivariant networks for tagging collisions in particle colliders only to break the symmetry by introducing dummy inputs that identify the collision axis. \citet{van2018learning} use the marginal likelihood to learn approximate invariances in Gaussian processes from data. In a related procedure, \citet{benton2020learning} learn the \emph{extent} of  symmetries in neural networks using the reparametrization trick and test time augmentation. While sharing some commonalities with RPP, this method is not aimed at achieving approximate equivariance and cannot bake equivariance into the model architecture.

A separate line of work has attempted to combine the extreme flexibility of the Vision Transformer (ViT) \citep{dosovitskiy2020image} with the better sample efficiency of convolutional networks, by incorporating convolutions at early layers \citep{xiao2021early} or making the self attention layer more like a convolution \citep{d2021convit,dai2021coatnet}. Most similar to our work, ConViT \citep{d2021convit} uses a gating mechanism for adding a soft locality bias to the self attention mechanism in Vision Transformers. ConViT and RPP share the same motivation, but while ConViT is designed specifically for biasing towards locality in the self attention layer, RPP is a general approach that we can apply broadly with other kinds of layers, symmetries, or architectural constraints.

Outside of equivariance, adding model outputs to a much more restrictive base model 
has been a fruitful idea employed in multiple contexts. The original ResNet \citep{he2016deep,he2016identity} drew on this motivation, with shortcut connections. \citet{johannink2019residual} and \citet{silver2018residual} proposed Residual Reinforcement Learning, whereby the RL problem is split into a user designed controller using engineering principles and a flexible neural network policy learned with RL. Similarly, in modeling dynamical systems, one approach is to incorporate a base parametric form informed by models from physics or biology, and only learn a neural network to fit the delta between the simple model and reality \citep{kashinath2021physics,liu2021physics}. 

There have been several works tackling symmetries and equivariance in RL, such as permutation equivariance for multi-agent RL \citep{sukhbaatar2016learning,jiang2018graph,liu2020pic}, as well exploring reflection symmetry for continuous control tasks \citep{abdolhosseini2019learning}, and discrete symmetries in the more general framework of MDP homomorphisms \citep{van2020mdp}. However, in each of these applications the symmetries need to be exact, and the complexities of real data often require violating those symmetries. Although not constructed with this purpose, some methods which use regularizers to enforce equivariance \citep{van2020plannable} could be used for approximate symmetries. Interestingly, the value of approximate symmetries of MDPs has been explored in some theoretical work \citep{ravindran2004approximate,taylor2008bounding}, but without architectures that can make use of it. Additionally, data augmentation, while not able to bake in architectural equivariance, has been successfully applied to encouraging equivariance on image tasks \citep{kostrikov2020image} and recently even on tabular state vectors \citep{lin2020invariant,mavalankar2020goal}.

\section{Background}
In order develop our method, we first review the concept of group symmetries, how representations formalize the way these symmetries act on different objects.

\paragraph{Group Symmetries}
In the machine learning context, a symmetry group $G$ can be understood as a set of invertible transformations under which an object is the same, such as reflections or rotations. These symmetries can act on many different kinds of objects. A rotation could act on a simple vector, a $2$d array like an image, a complex collection objects like the state space of a robot, or more abstractly on an entire classification problem or Markov Decision Process (MDP).

\paragraph{Representations}
The way that symmetries act on objects is described by a \emph{representation}.
Given an object in an $n$-dimensional vector space $V$, a group representation is a mapping $\rho: G\rightarrow \mathbb{R}^{n\times n}$, yielding a matrix which acts on $V$. Vectors $v\in V$ are transformed $v \mapsto \rho(g)v$. In deep learning, each of the inputs and outputs to our models can be embedded in some vector space: an $m\times m$ sized $\mathrm{rgb}$ image exists in $\mathbb{R}^{3m^2}$, and a node valued function on a graph of $m$ elements exists within $\mathbb{R}^{m}$. The representation $\rho$ specifies how each of these objects transform under the symmetry group $G$.

These representations can be composed of multiple simpler subrepresentations, describing how each object within a collection transforms. For example given the representation $\rho_1$ of rotations acting on a vector in $\mathbb{R}^3$, and a representation $\rho_2$ of how rotations act on a $3\times 3$ matrix, the two objects concatenated together have a representation given by $\rho_1(g)\oplus \rho_2(g) = \begin{bmatrix} \rho_1(g) & 0 \\ 0  & \rho_2(g) \\ \end{bmatrix}$, where the two matrices are concatenated along the diagonal. Practically this means we can represent intricate and multifaceted structures by breaking them down into their component parts and defining how each part transforms. For example, we may know that the velocity vector, an orientation quaternion, a joint angle, and a control torque all transform in different ways under a left-right reflection, and one can accommodate this information into the representation.

\paragraph{Equivariance}
Given some data $X$ with representation $\rho_{\mathrm{in}}$, and $Y$ with representation $\rho_{\mathrm{out}}$, we may wish to learn some mapping $f: X \to Y$.
A model $f$ is equivariant \citep{cohen2016group}, if applying the symmetry transformation to the input is equivalent to applying it to the output
\begin{equation*}
    f(\rho_{\mathrm{in}}(g)x) = \rho_{\mathrm{out}}(g) f(x).
\end{equation*}
In other words, it is not the symmetry of $X$ or $Y$ that is relevant, but the symmetry of the function $f$ mapping from $X$ to $Y$. If the true relationship in the data has a symmetry, then constraining the hypothesis space to functions $f$ that also have the symmetry makes learning easier and improves generalization \citep{elesedy2021provably}. Equivariant models have been developed for a wide variety of symmetries and data types like images \citep{cohen2016group,worrall2017harmonic,zhou2017oriented,weiler2019general}, sets \citep{zaheer2017deep,maron2020learning}, graphs \citep{maron2018invariant}, point clouds \citep{anderson2019cormorant,fuchs2020se,satorras2021n}, dynamical systems \citep{finzi2020generalizing}, jets \citep{bogatskiy2020lorentz}, and other objects \citep{wang2020equivariant,finzi2021practical}.

\section{Residual Pathway Priors} \label{sec: methods}
In this section, we introduce Residual Pathway Priors (RPPs). The core implementation of the RPP is to expand each layer in model into a sum of both a restrictive layer that encodes the hard architectural constraints and a generic more flexible layer, but penalize the more flexible path via a lower prior probability. Through the difference in prior probability, explanations of the data using only the constrained solutions are prioritized by the model; however, if the data are more complex the residual between the target and the constrained layer will be explained using the flexible layer. We can apply this procedure to any restriction priors, such as linearity, locality, Markovian structure, and, of course, equivariance.

The Residual Pathway Prior draws inspiration from the residual connections in ResNets \citep{he2016deep,he2016identity}, whereby training stability and generalization improves by providing multiple paths for gradients to flow through the network that have different properties.
One way of interpreting a residual block and shortcut connection $f(x) = x + h(x)$ in combination with l2 regularization, either explicitly from weight decay or implicitly from the training dynamics \citep{neyshabur2014search}, is as a prior that places higher prior likelihood on the much simpler identity mapping than on the more flexible function $h(x)$. In this way, $h(x)$ need only explain the the difference between what is explained in the previous layer (passed through by $I$) and the target.

We can leverage a similar approach to convert the hard constraints of a restriction prior specified through a given network architecture into a soft prior that merely places higher prior density on such models. Supposing we have an equivariant layer $A(x)$, and a more flexible non-equivariant layer $B(x)$ which contains $A$ as a special case, we can allow the model to explain part of the data with $A$, and part with $B$, by forming the sum $A(x)+B(x)$. Given a prior on the size of $A$ and $B$, such as $p(A) \propto \exp{(-\|A\|^2/2\sigma_a^2)}$, and $p(B) \propto \exp{(- \|B\|^2/2\sigma_b^2)}$ with $\sigma_a>\sigma_b$, a MAP optimized model will favor explanations of the data using the more structured layer $A$, and only resort to using layer $B$ to explain the \emph{difference} between the target and what is already explained by the more structured model $A$. Adding these non-equivariant residual pathways to each layer of an equivariant model, we have a model that has the same expressivity of a network formed entirely of $B$ layers, but with the inductive bias towards a model formed entirely with the equivariant $A$ layers. We term this model a \emph{Residual Pathway Prior}.

To make the approach concrete, we first consider constructing equivariance priors using the constraint solving approach known as Equivariant Multi-Layer Perceptrons (EMLP) from  \citet{finzi2021practical}.

\paragraph{Equivariant MLPs}

EMLPs provide a method for automatically constructing exactly equivariant layers for any given group and representation by solving a set of constraints. The way in which the vectors are equivariant is given by a formal specification of the types of the input and output through defining their representations. Given some input vector space $V_\mathrm{in}$ with representation $\rho_\mathrm{in}$ and some output space $V_\mathrm{out}$ with representation $\rho_\mathrm{out}$ the space of all equivariant linear layers mapping $V_\mathrm{in}\rightarrow V_\mathrm{out}$ satisfies
\begin{equation*}
    \forall g \in G: \ \ \rho_\mathrm{out}(g)W = W\rho_\mathrm{in}(g).
\end{equation*}
 These solutions to the constraint form a subspace of matrices 
$\mathbb{R}^{n_\mathrm{out}\times n_\mathrm{in}}$ which can be solved for and described by a $r$ dimensional orthonormal basis $Q \in \mathbb{R}^{n_\mathrm{out}n_\mathrm{in}\times r}$. Linear layers can then be parametrized in this equivariant basis. The elements of $W$ can be parametrized $\mathrm{vec}(W)=Q\beta$ for $\beta \in \mathbb{R}^r$ for the linear layer $v\mapsto Wv$, and symmetric biases can be parametrized similarly.

\paragraph{Equivariance Priors with EMLP} In order to convert the hard equivariance constraints in EMLP into a soft prior over equivariance that can accommodate approximate symmetries, we can apply the RPP procedure from above to each these linear layers in the network. Instead of parametrizing the weights $W$ directly in the equivariant basis $\mathrm{vec}(W)=Q\beta$, we can instead define $W$ as the sum $W = A + B$ of an equivariant weight matrix $\mathrm{vec}(A)=Q\beta$ with a Gaussian prior over $\beta$ and an unconstrained weight matrix $B\sim \mathcal{N}(0,\sigma_b^2 I)$ with a Gaussian prior. A Gaussian prior over $\beta\sim \mathcal{N}(0,\sigma_a^2I)$ is equivalent to the Gaussian prior $A\sim \mathcal{N}(0,\sigma_a^2 QQ^\top)$. Since $Q$ is an orthogonal matrix, we can breakdown the covariance of $B$ into the covariance in the equivariant subspace $Q$ as well as the covariance in the orthogonal complement $P$, $\sigma_b^2I = \sigma_b^2QQ^\top+\sigma_b^2PP^\top$. Therefore the sum of the weight matrices is distributed as
$ A+B = W \sim \mathcal{N}(0,(\sigma_a^2+\sigma_b^2) QQ^\top + \sigma_b^2PP^\top)$.

\emph{Regardless} of the values of the prior variances $\sigma_a^2$ and $\sigma_b^2$, solutions in the equivariant subspace $QQ^\top$ are automatically favored by the model and assigned higher prior probability mass than those in the subspace $PP^\top$ that violate the constraint. Even if $\sigma_b>\sigma_a$, the model still favors equivariance because the equivariance solutions are contained in the more flexible layer $A$. We show in Section \ref{sec: prior-var} that RPPs are insensitive to the choice of $\sigma_a$ and $\sigma_b$, provided that $\sigma_a$ is large enough to be able to fit the data.
By replacing each of the equivariant linear layers in an EMLP with a sum of an equivariant layer and an unconstrained layer and adding in the negative prior likelihood to the loss function, we produce an RPP-EMLP that can accommodate approximate or incorrectly specified symmetries. \footnote{For the EMLP that uses gated nonlinearities which do not always reduce to a standard Swish, we likewise add a more general Swish weighted by a parameter with prior variance $\sigma_b^2$.}

\paragraph{RPPs With Other Equivariant Models}

While in EMLP equivariant bases are solved for explicitly, the RPP can be applied to the linear layers in other equivariant networks in precisely the same way. A good example is the translationally equivariant convolutional neural network (CNN), which can be viewed as a restricted subset of a fully connected network. Though the layers are parametrized as convolutions, the convolution operation can be expressed as a Toeplitz matrix residing within the space of dense matrices. Adding the convolution to a fully connected layer and choosing a prior variance $\sigma_a^2$ and $\sigma_b^2$ over each,
we have the same RPP prior
\begin{equation}\label{eqn:rpp-conv}
    W \sim N(0, \sigma_a^2 QQ^\top + \sigma_b^2 I)
\end{equation}
where $Q$ is the basis of (bi-)Toeplitz matrices corresponding to $3\times 3$ filters. This RPP CNN has the biases of convolution but can readily fit non translationally equivariant data. We can similarly create priors with the biases of other equivariant models like GCNNs \citep{cohen2016group}, without any hard constraints. We can even apply the RPP principle to the breaking of a given symmetry group to a subgroup.

\section{How and Why RPPs Work}\label{sec: howandwhy}

We explore how and why RPPs work on a variety of domains, applying RPPs where (1) constraints are known to be helpful, (2) cannot fully describe the problem, and (3) are misspecified.

\subsection{Dynamical Systems and Levels of Equivariance}\label{sec: dynamical-systems}

In order to better understand how and why residual pathway priors interact with the symmetries of the problem we move to settings in which we can directly control both the type of symmetry and the level to which the symmetries are violated. We examine how RPPs coupled with EMLP networks (RPP-EMLP) perform on the inertia and double pendulum datasets featured in \citet{finzi2021practical} in $3$ experimental settings: $(i)$ the original inertia and double pendulum datasets which preserve exact symmetries with respect to the to $\mathrm{O}(3)$ and $\mathrm{O}(2)$ groups respectively; $(ii)$ modified versions of these datasets with additional factors (such as wind on the double pendulum) that lead to approximate symmetries; and $(iii)$ versions with misspecified symmetry groups that break the symmetries entirely (described in \autoref{app:experimental-details}). 

The results for these $3$ settings are given in Figure \ref{fig:dynamics}. Across all settings RPP-EMLP match the performance of EMLP when symmetries are exact, perform as well as an MLP when the symmetry is misspecified and better than both when the symmetry is approximate. For these experiments we use a prior variance of $\sigma_a^2 = 10^5$ on the EMLP weights and $\sigma_b^2=1$ on the MLP weights.

\begin{figure}
\begin{subfigure}{.33\textwidth}
  \centering
  \includegraphics[width=\linewidth]{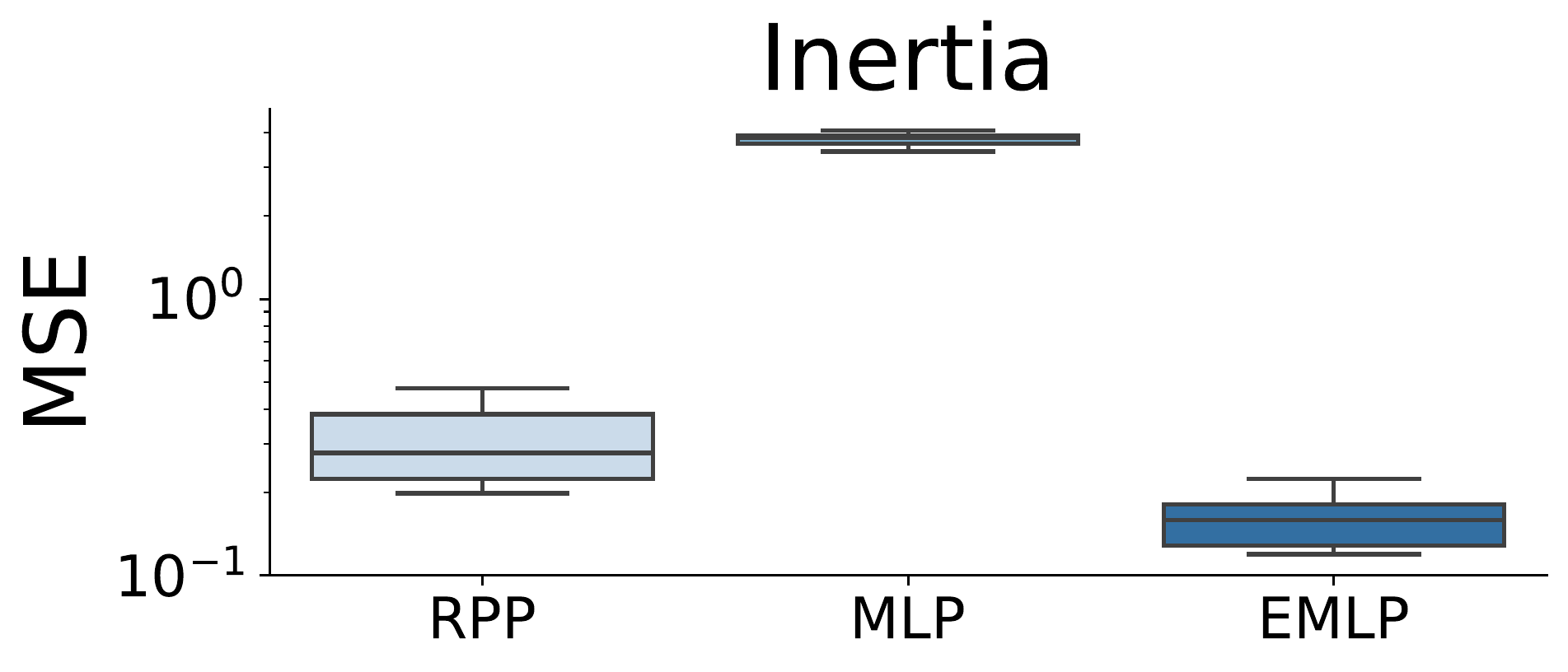}  
\end{subfigure}
\begin{subfigure}{.33\textwidth}
  \centering
  \includegraphics[width=\linewidth]{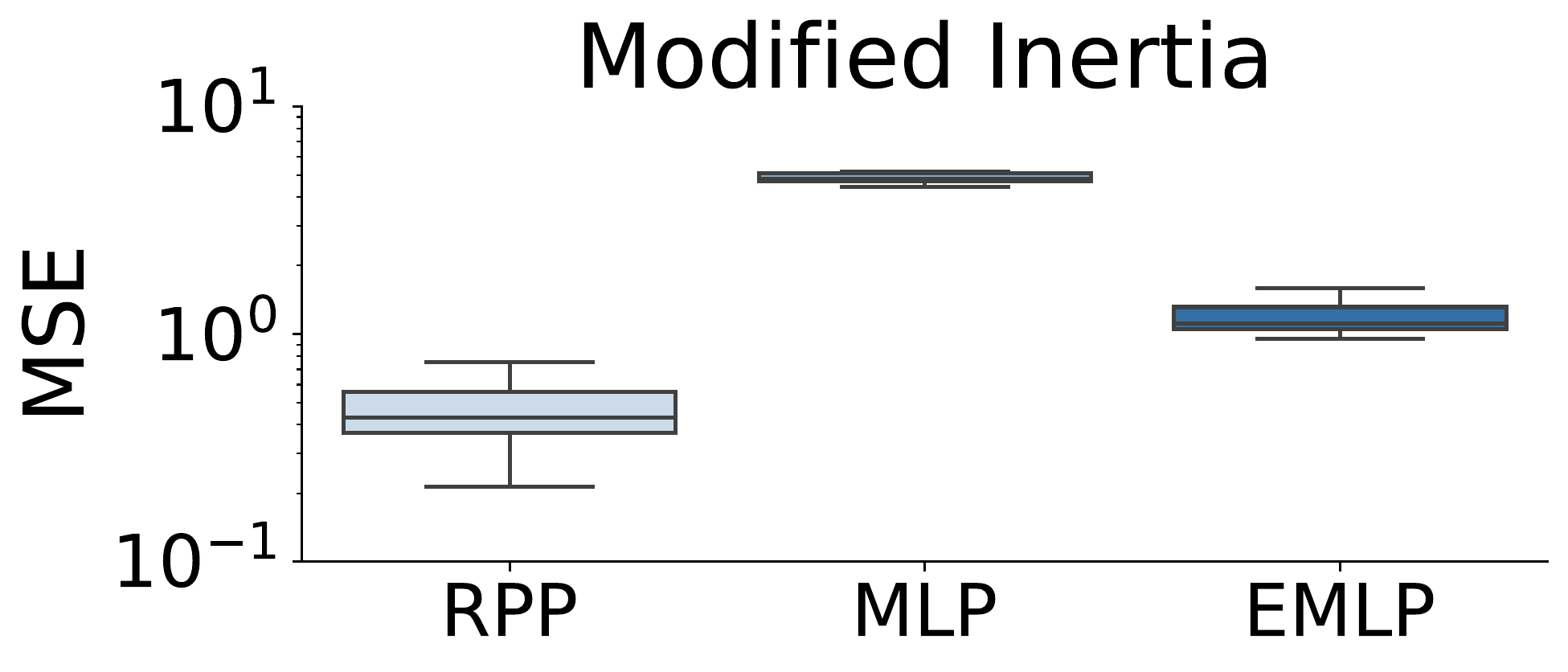}  
\end{subfigure}
\begin{subfigure}{.33\textwidth}
  \centering
  \includegraphics[width=\linewidth]{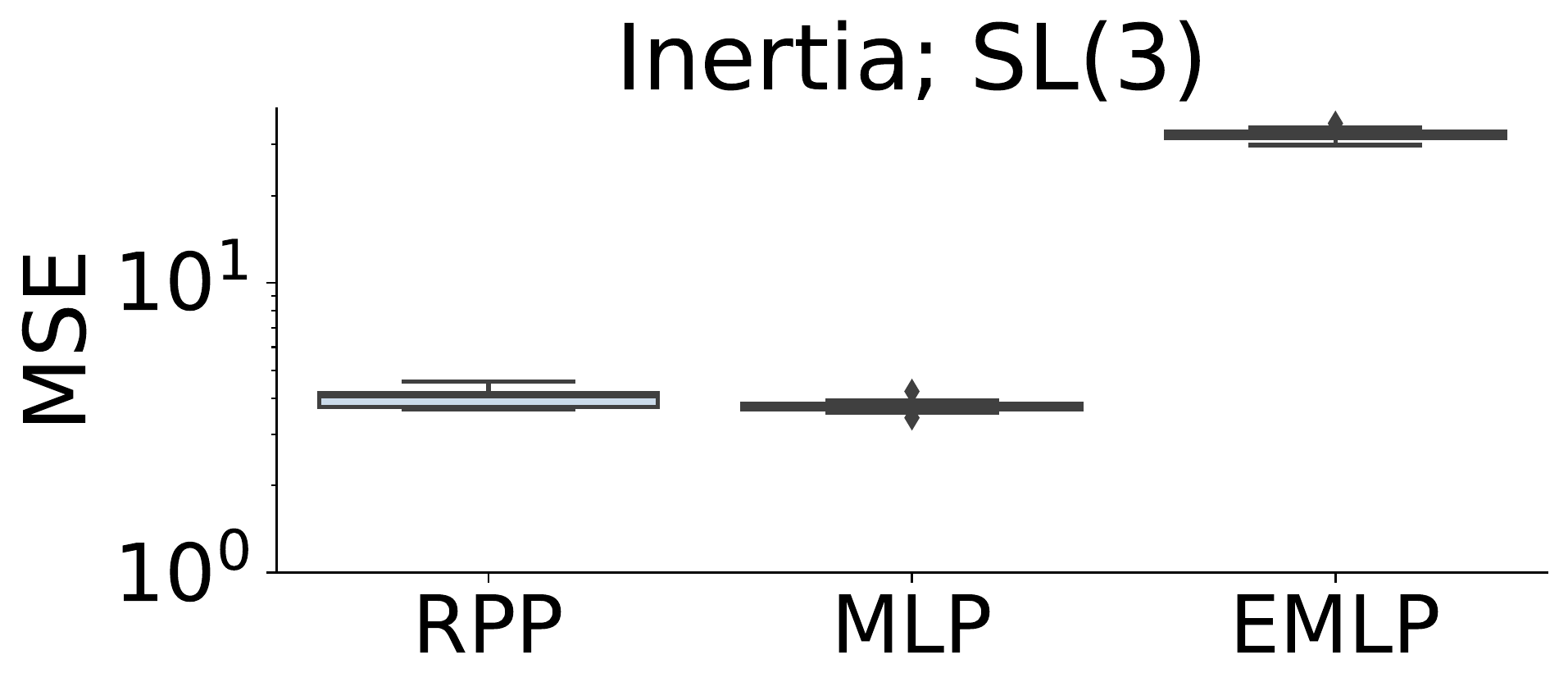}  
\end{subfigure}

\begin{subfigure}{.33\textwidth}
  \centering
  \includegraphics[width=\linewidth]{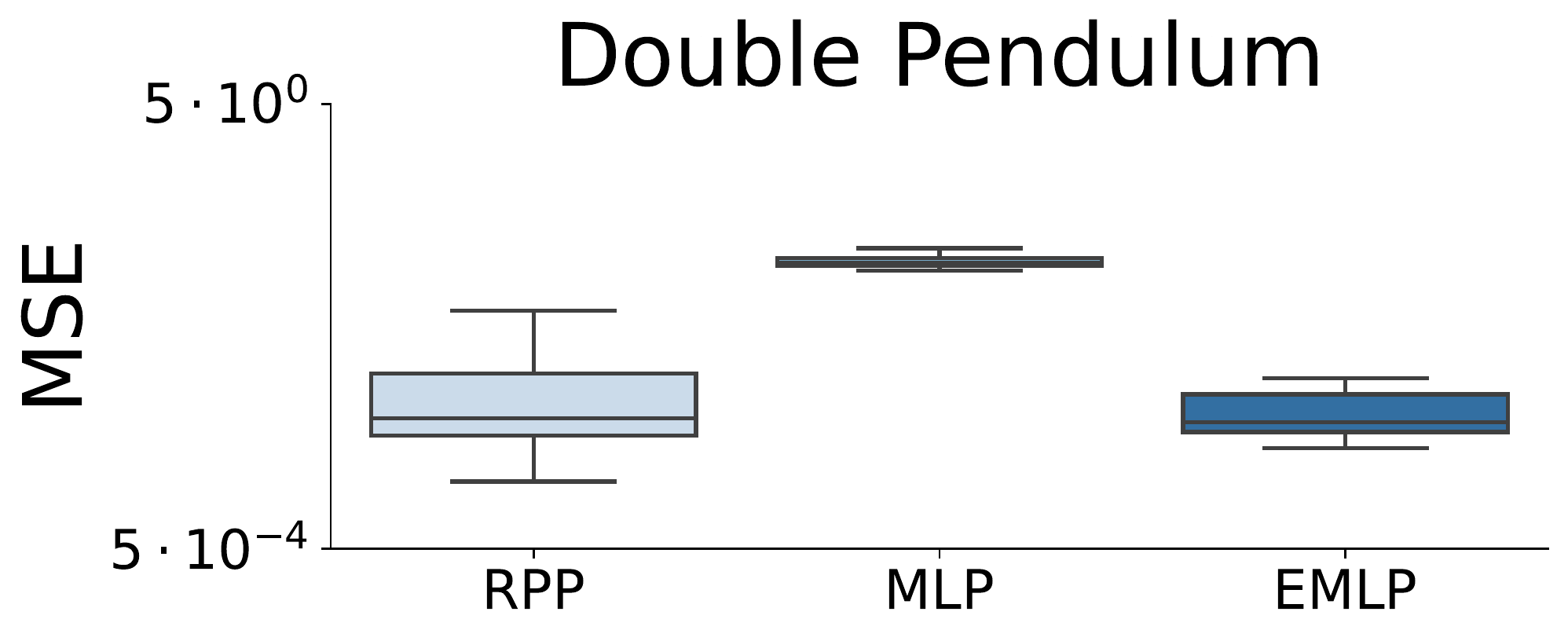}  
  \caption{Exact Symmetries}
  \label{fig:perfect-sym}
\end{subfigure}
\begin{subfigure}{.33\textwidth}
  \centering
  \includegraphics[width=\linewidth]{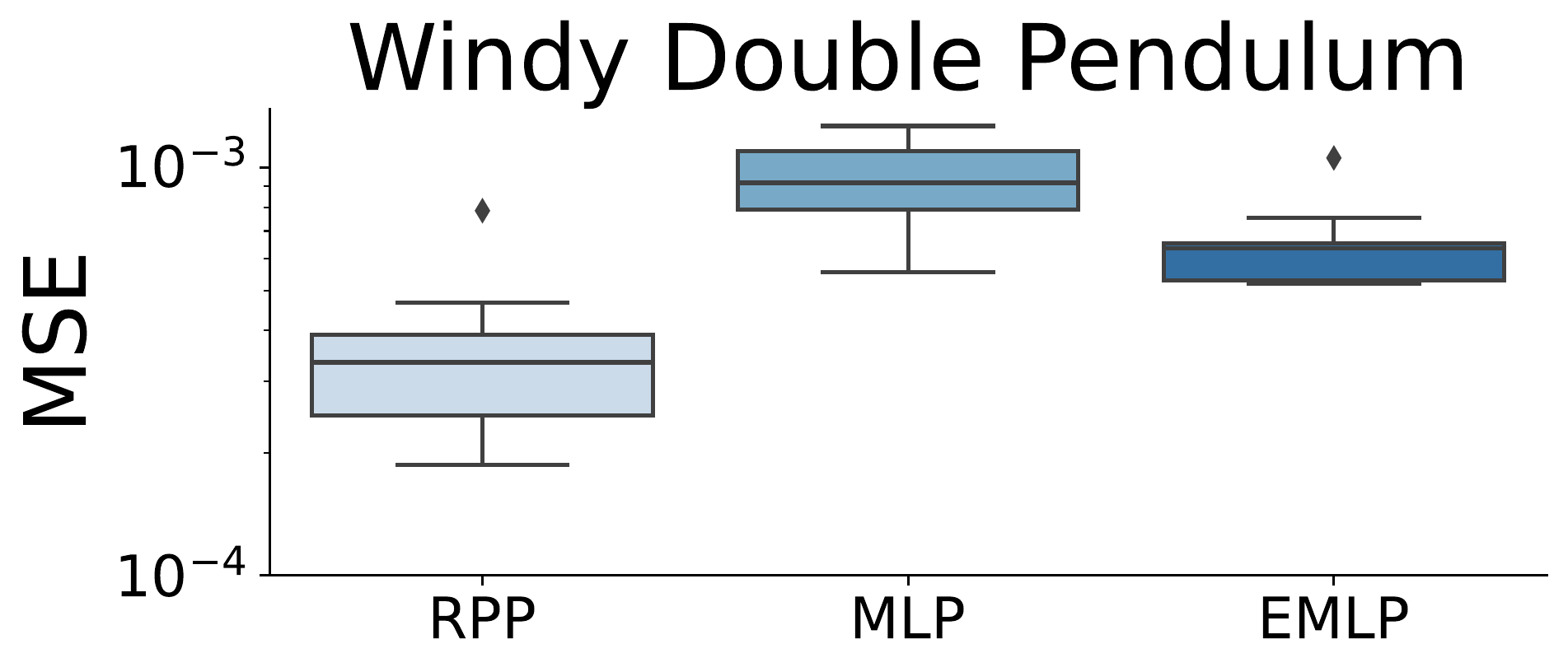}  
  \caption{Approximate Symmetries}
  \label{fig:partial-sym}
\end{subfigure}
\begin{subfigure}{.33\textwidth}
  \centering
  \includegraphics[width=\linewidth]{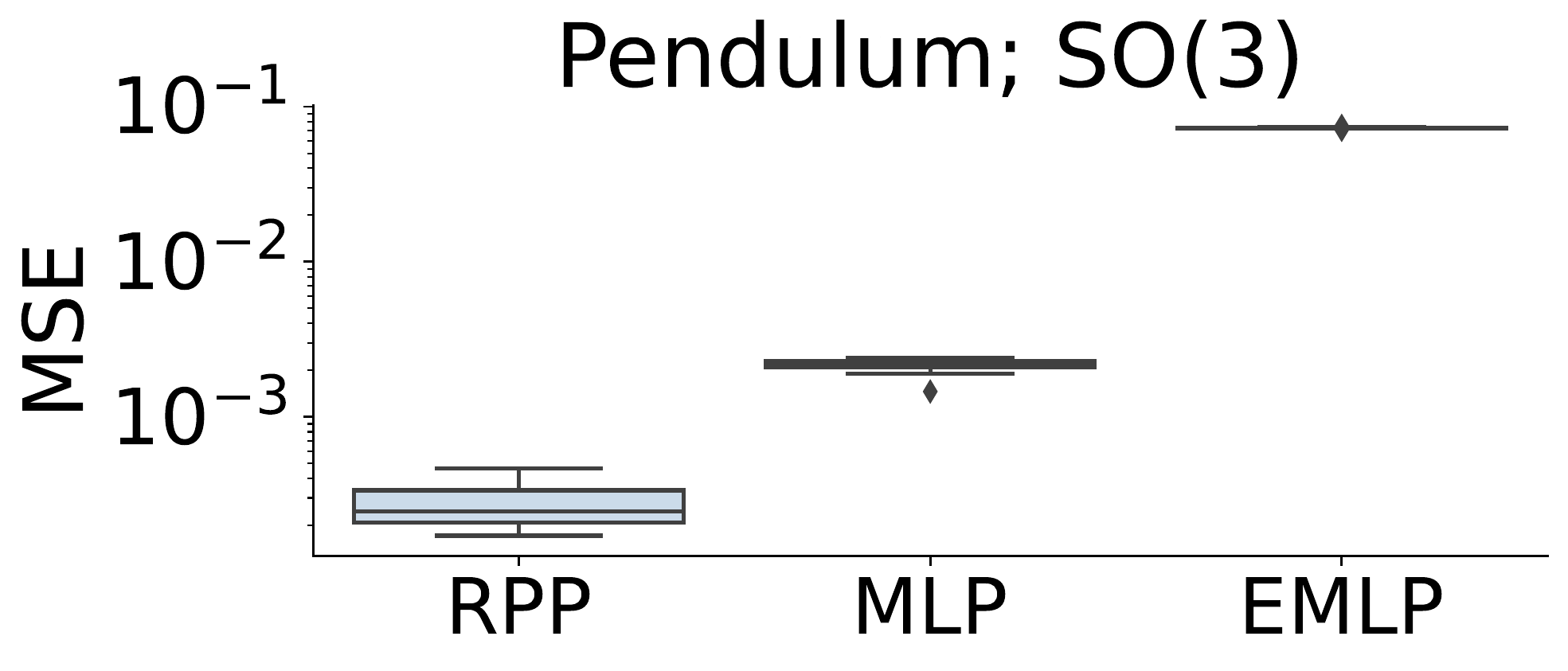}  
  \caption{Mis-specified Symmetries}
  \label{fig:misspec-sym}
\end{subfigure}
\caption{A comparison of test performance over $10$ independent trials using RPP-EMLP and equivalent EMLP and MLP models on the inertia (\textbf{top}) and double pendulum (\textbf{bottom}) datasets in which we have three varying levels of symmetries. The boxes represent the interquartile range, and the whiskers the remainder of the distribution. \textbf{Left:} perfect symmetries in which EMLP and the equivariant components of RPP-EMLP exactly capture the symmetries in the data. \textbf{Center:} approximate symmetries in which the perfectly symmetric systems have been modified to include some non-equivariant components. \textbf{Right:} mis-specified symmetries in which the symmetric components of EMLP and RPP-EMLP do not reflect the symmetries present in the data.}
\label{fig:dynamics}
\end{figure}

\paragraph{Exact Symmetries} 

As part of the motivation, RPPs should properly allocate prior mass to both constrained and unconstrained solutions, we test cases in which symmetries are exact, and show that RPP-EMLP is capable of performing on par with EMLP which only admits solutions with perfect symmetry. 
The results in Figure \ref{fig:perfect-sym} show that although the prior over models as described RPP-EMLP is broader than that of EMLP (as we can admit non-equivariant solutions) in the presence of perfectly equivariant data RPP-EMLP do not hinder performance, and we are able to generalize nearly as well as the perfectly prescribed EMLP model.

\paragraph{Approximate symmetries}

To better showcase the ideas of Figure \ref{fig:probability} we compare RPP-EMLPs to EMLPs and MLPs on the modified inertia and windy pendulum datasets. In these datasets we can think about the systems as primarily equivariant but containing non-equivariant contributions. As shown in \ref{fig:partial-sym} these problems are best suited for RPP-EMLP as MLPs have no bias towards the approximately symmetry present in the data, and EMLPs are overly constrained in this setting.

\paragraph{Misspecified symmetries}
In contrast to working with perfect symmetries and showing that RPP-EMLPs are competitive with EMLPs, we also show that when symmetries are \emph{mis}specified the bias towards equivariant solutions does not hinder the performance of RPP-EMLPs. For the inertia dataset we substitute the group equivariance in EMLP from $\mathrm{O}(3)$ to the overly large group $\mathrm{SL}(3)$ consisting of all volume and orientation preserving linear transformations, not just the orthogonal ones. For the double pendulum dataset, we substitute $\mathrm{O}(2)$ symmetry acting on $\mathbb{R}^3$ with the larger $\mathrm{SO}(3)$ rotation group that contains it but is not a symmetry of the dataset.  

By purposefully misspecifying the symmetry in these datasets we intentionally construct EMLP and RPP-EMLP models with incorrect inductive biases. In this setting EMLP is incapable of making accurate predictions as it has a hard constraint on an incorrect symmetry. 
Figure \ref{fig:misspec-sym} shows that even in cases where the model is intentionally mis-specified that RPPs can overcome a poorly aligned inductive bias and recover solutions that perform as well as standard MLPs, even where EMLPs fail.

\subsection{Prior Levels of Equivariance}\label{sec: prior-var}
\begin{figure}
\begin{subfigure}{.5\textwidth}
    \centering
        \includegraphics[width=\linewidth]{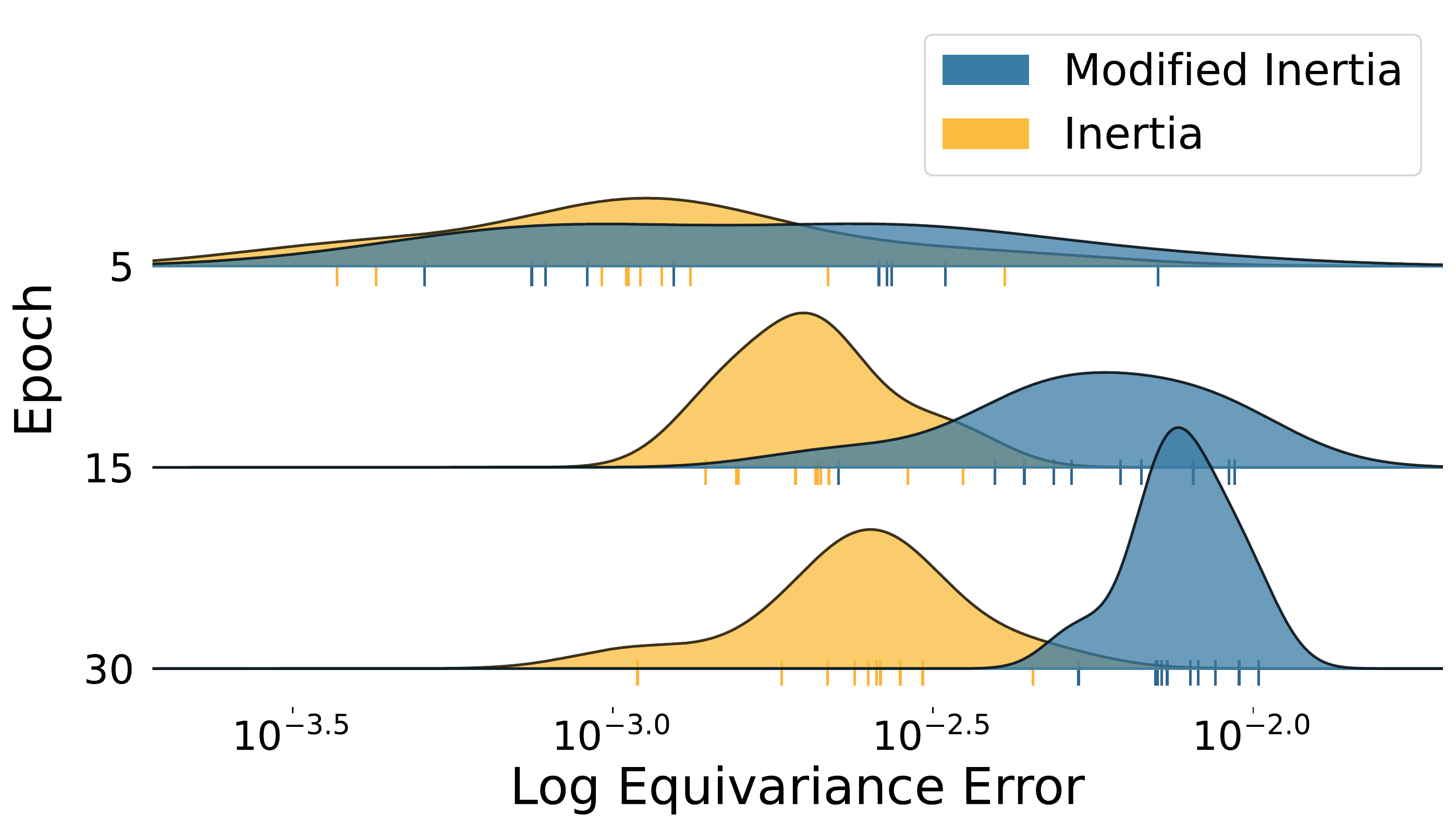}
    \end{subfigure}
    \begin{subfigure}{.5\textwidth}
      \centering
    \includegraphics[width=\linewidth]{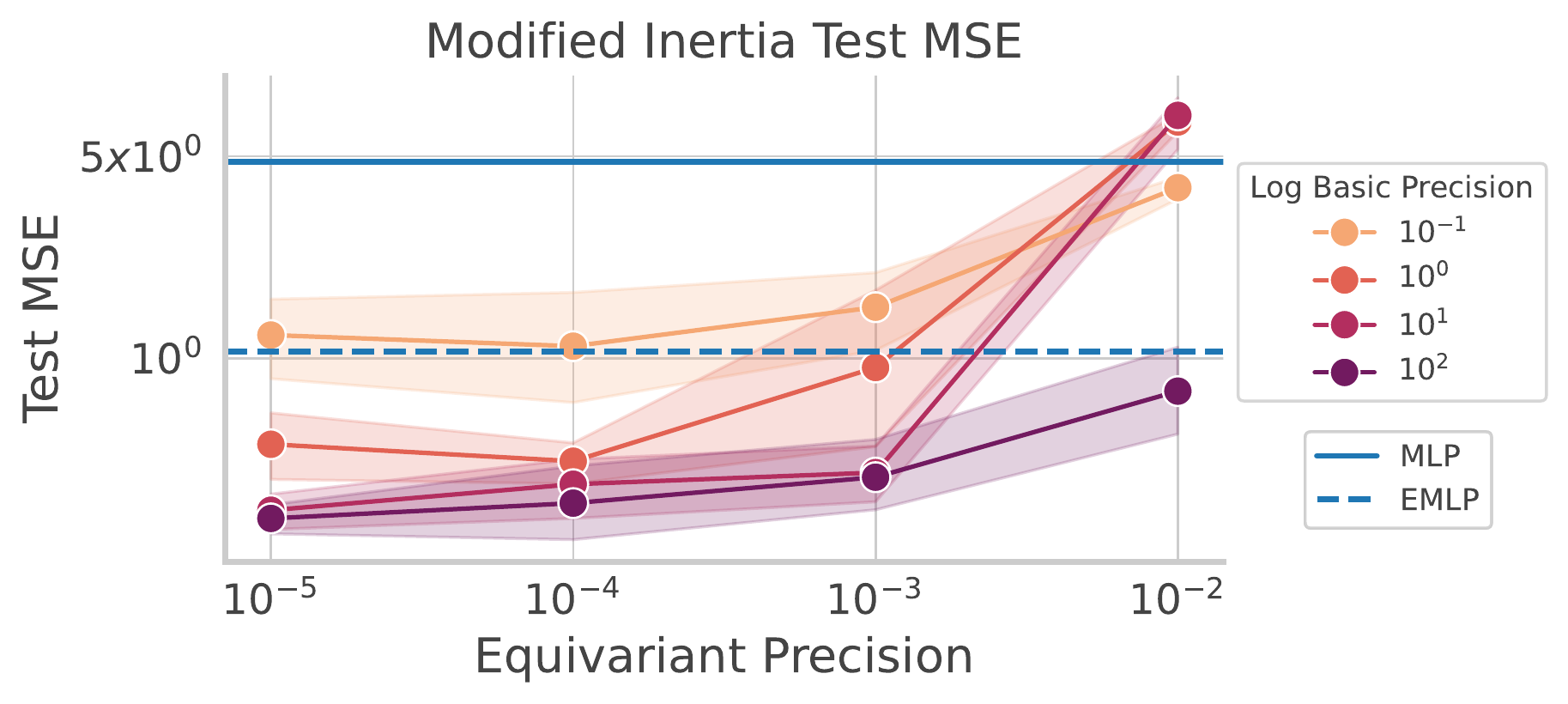}
    %\label{fig:post-equiv}
    \end{subfigure}
    
    \caption{\textbf{Left:} Kernel density estimators of log equivariance error across training epochs for $10$ independently trained networks. 
    Here the color denotes the dataset these models were trained on. Treating these samples as a proxy for posterior density, we see that on the non-equivariant Modified Inertia dataset, the posterior is shifted upward to match the level of equivariance in the data during training.
    \textbf{Right:} Test MSE as a function of the weight decay parameters on the equivariant and basic weights on the modified inertia dataset. We observe that so long as the prior in the basis of equivariant weights is broad enough, we can achieve low test error with RPPs.}
    \label{fig: prior-var-ablation-post-equiv}
\end{figure}

To test the effect of prior variances we use the modified inertia dataset, which represents a version of a problem in which perfect equivariance has been broken by adding new external forces to the dynamical system.
Shown in Figure \ref{fig: prior-var-ablation-post-equiv} (right) is a comparison of mean squared error on test data as a function of the prior precision terms on both the equivariance and basic weights. As a general trend we see that when the regularization on the equivariant weights is too high (equivalent to a concentrated prior around $0$) we find instability in test performance, yet when we apply a broad prior to the equivariant weights performance is typically both better in terms of MSE, and more stable to the choice of prior on the basic model weights. 

As the prior variances over the equivariant basis $Q$ and the non-equivariant basis $P$ describe our bias towards or away from equivariant solutions we investigate how the choice of prior variance relates to the level of symmetry present in a given dataset. In the windy pendulum dataset we have control over the level of wind and thus how far our system is from perfect equivariance.

\subsection{Posterior Levels of Equivariance}\label{sec: post-equiv}

RPPs describe a method for setting a prior over equivariance, and in the presence of new data we expect the posterior distribution over equivariance to change accordingly. 
Using samples from a deep ensemble to query points of high density in the posterior we estimate how the distribution over equivariance error progresses through training. Recalling that with an equivariant function $f$ we have $\rho_2(g) f(x) = f(\rho_1(g) x)$, we compute equivariance error as 
\begin{equation}\label{eqn:equiv-error}
    \mathrm{EquivErr}(f, x) = \mathrm{RelErr}(\rho_2(g) f(x),f(\rho_1(g)x)) \ \ \mathrm{where} \ \ \mathrm{RelErr}(a,b) = \frac{\|a-b\|}{\|a\|+\|b\|}.
\end{equation}

We train one deep ensemble on the inertia dataset which exhibits perfect symmetry, and another on the modified inertia dataset which has only partial symmetry, with each deep ensemble being comprised of $10$ individual models using the same procedure as in Section \ref{sec: dynamical-systems}.
In Figure \ref{fig: prior-var-ablation-post-equiv} (left) we see that throughout training the models trained on the modified inertia concentrate around solutions with substantially higher equivariance error than models trained on the dataset with the exact symmetry. This figure demonstrates one of the core desiderata of RPPs: that we are able to converge to solutions with an appropriate level of equivariance for the data.

\subsection{RPPs and Convolutional Structure}\label{sec: rpp-conv}

Using the RPP-Conv specified by the prior in Eqn \ref{eqn:rpp-conv} we apply the model to CIFAR-$10$ classification and UCI regression tasks where the inputs are reshaped to zero-padded two dimensional arrays and treated as images. Notably, the model is still an MLP and merely has larger prior variance in the convolutional subspace. As a result it can perform well on image datasets where the inductive bias is aligned, as well as on the UCI data despite not being an image dataset as shown in Table \ref{table:rppconv}. While retaining the flexibility of an MLP, the RPP performs better than the locally connected MLPs trained with $\beta$-lasso in \citet{neyshabur2020towards} which get $14\%$ error on CIFAR-$10$. The full details for the architectures and training procedure are given in Appendix \ref{app:experimental-details}.

\begin{table}[h]
    \centering
    \begin{tabular}{cccc ccc}
        \toprule
          & CIFAR-$10$ & Energy & Fertility & Pendulum & Wine\\
         \midrule
         MLP & $37.61 \pm 0.56$ & $0.39 \pm 0.48$ & $0.049 \pm 0.0044$ & $4.65 \pm 0.50$ & $0.66 \pm 0.058$\\
         RPP & $12.62 \pm 0.34$ & $0.73 \pm 0.44$ & $0.060 \pm 0.0097$ & $4.25 \pm 0.50$ & $0.69 \pm 0.031$\\
         Conv& $12.03 \pm 0.46$ & $1.34 \pm 0.38$ & $0.076 \pm 0.0157$ & $4.63 \pm 0.36$ & $0.79 \pm 0.092$\\
         \bottomrule
    \end{tabular}
    \vspace{2mm}
    \caption{Mean test classification error on CIFAR-$10$ and MSE on $4$ UCI regression tasks, with one standard deviation errors taken over $10$ trials. Similar to Figure \ref{fig:dynamics}, we find that whether the constrained convolutional structure is helpful (CIFAR) or not (UCI), RPP-Conv performs similarly to the model with the correct level of complexity.}
    \label{table:rppconv}
\end{table}

\section{Approximate Symmetries in Reinforcement Learning}\label{sec: rl-priors}
Both model free and model based reinforcement learning present opportunities to take advantage of structure in the data for predictive power and data efficiency. On the one hand stands the use of model predictive control in the engineering community where finely specified dynamics models are constructed by engineers and only a small number of parameters are fit with system identification to determine mass, inertia, joint stiffness, etc. On the other side of things stands the hands off approach taken in the RL community, where general and unstructured neural networks are used for both transition models \citep{chua2018deep,wang2019exploring,janner2019trust} as well as policies and value functions \citep{haarnoja2018soft}. The state and action spaces for these systems are highly complex with many diverse inputs like quaternions, joint angles, forces, torques that each transform in different ways under a symmetry transformation like a left-right reflection or a rotation. As a result, most RL methods treat these spaces a black box ignoring all of this structure, and as a result they tend to require tremendous amounts of training data, making it difficult to apply to real systems without the use of simulators.

\begin{figure}
\centering
\begin{subfigure}[b]{.23\textwidth}
  \centering
  \includegraphics[width=\linewidth]{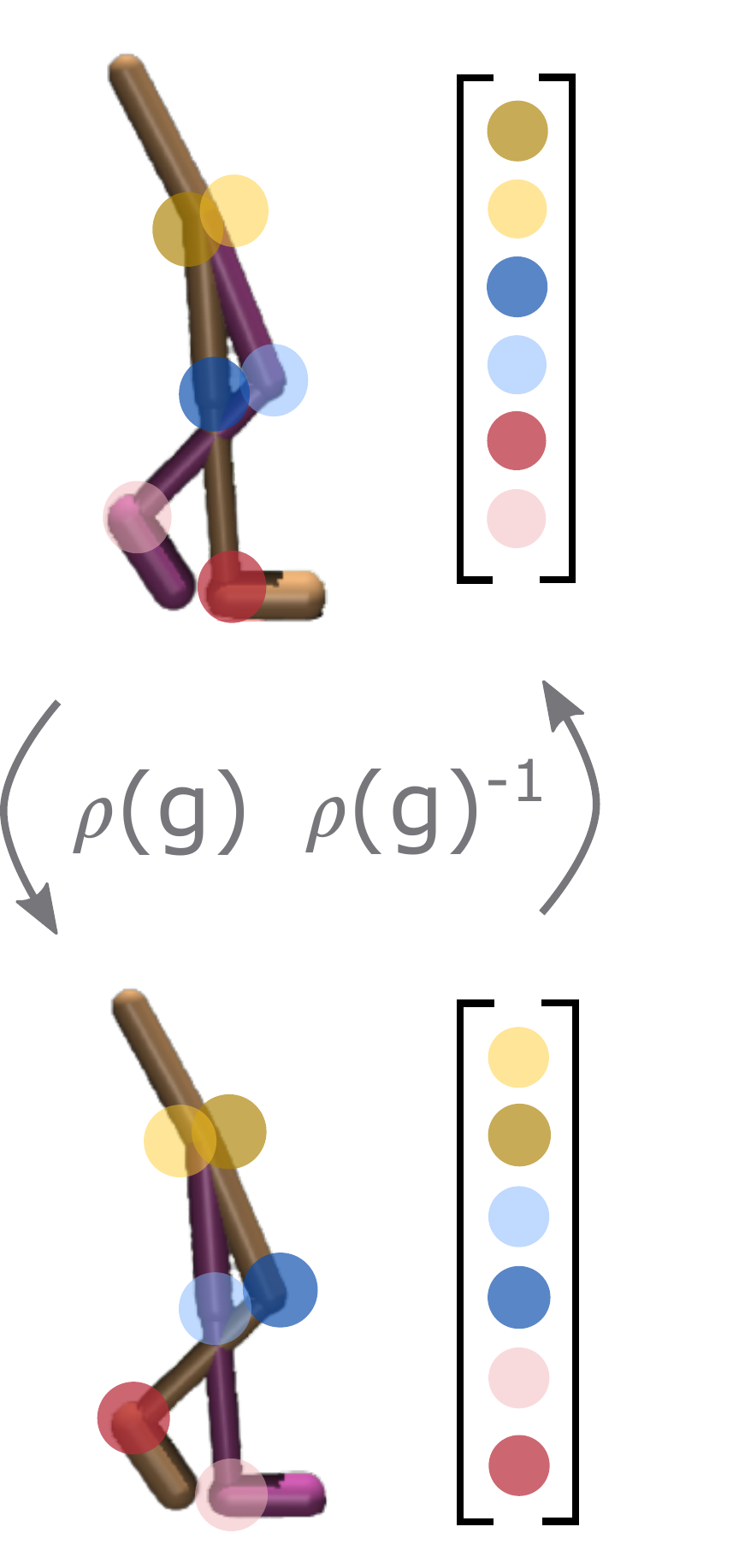}
  \caption{Walker2d Left-Right}
  \label{fig:perfect-sym}
\end{subfigure}
\begin{subfigure}[b]{.23\textwidth}
  \centering
  \includegraphics[width=\linewidth]{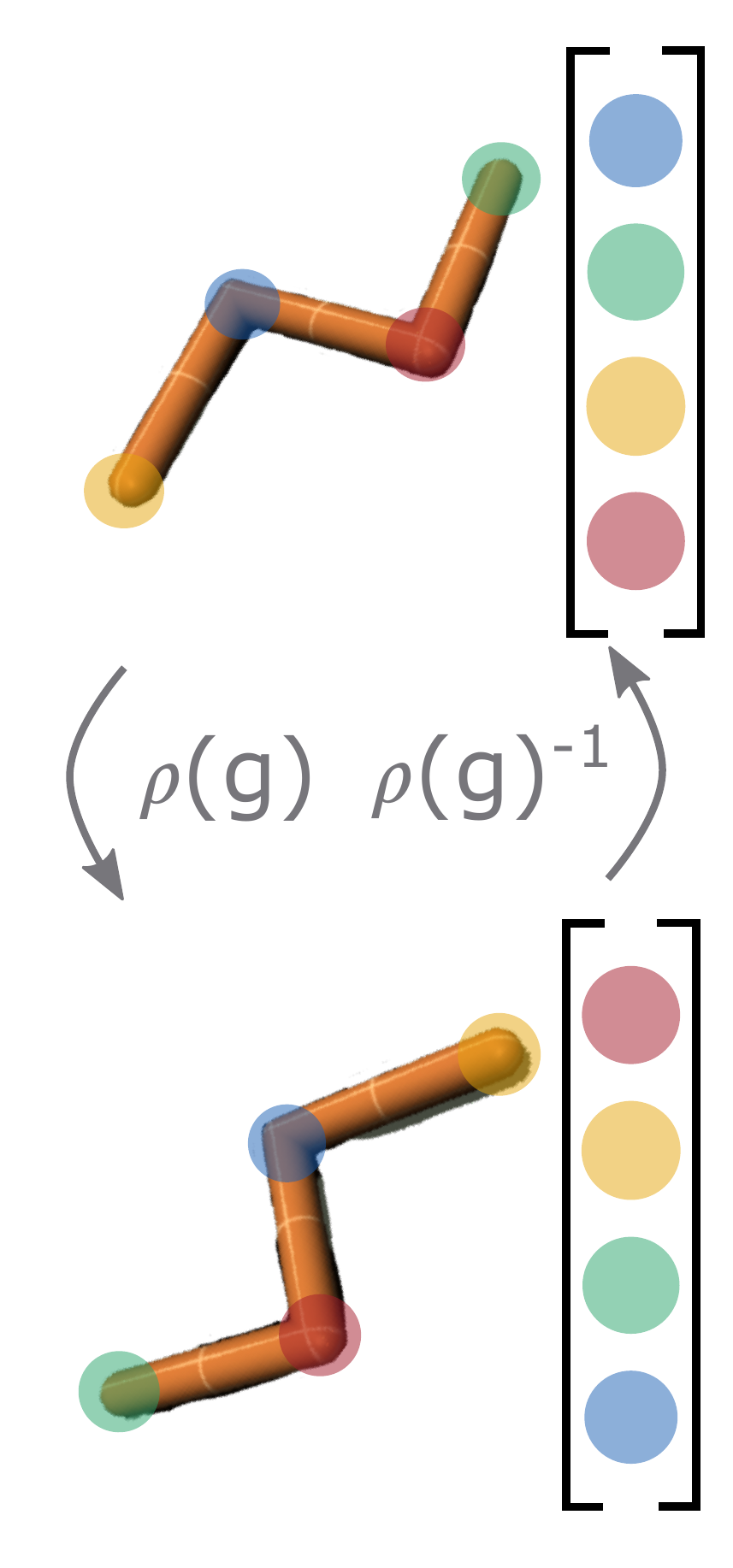}  
  \caption{Swimmer Front-Back}
  \label{fig:partial-sym}
\end{subfigure}
\begin{subfigure}[b]{.23\textwidth}
  \centering
  \includegraphics[width=\linewidth]{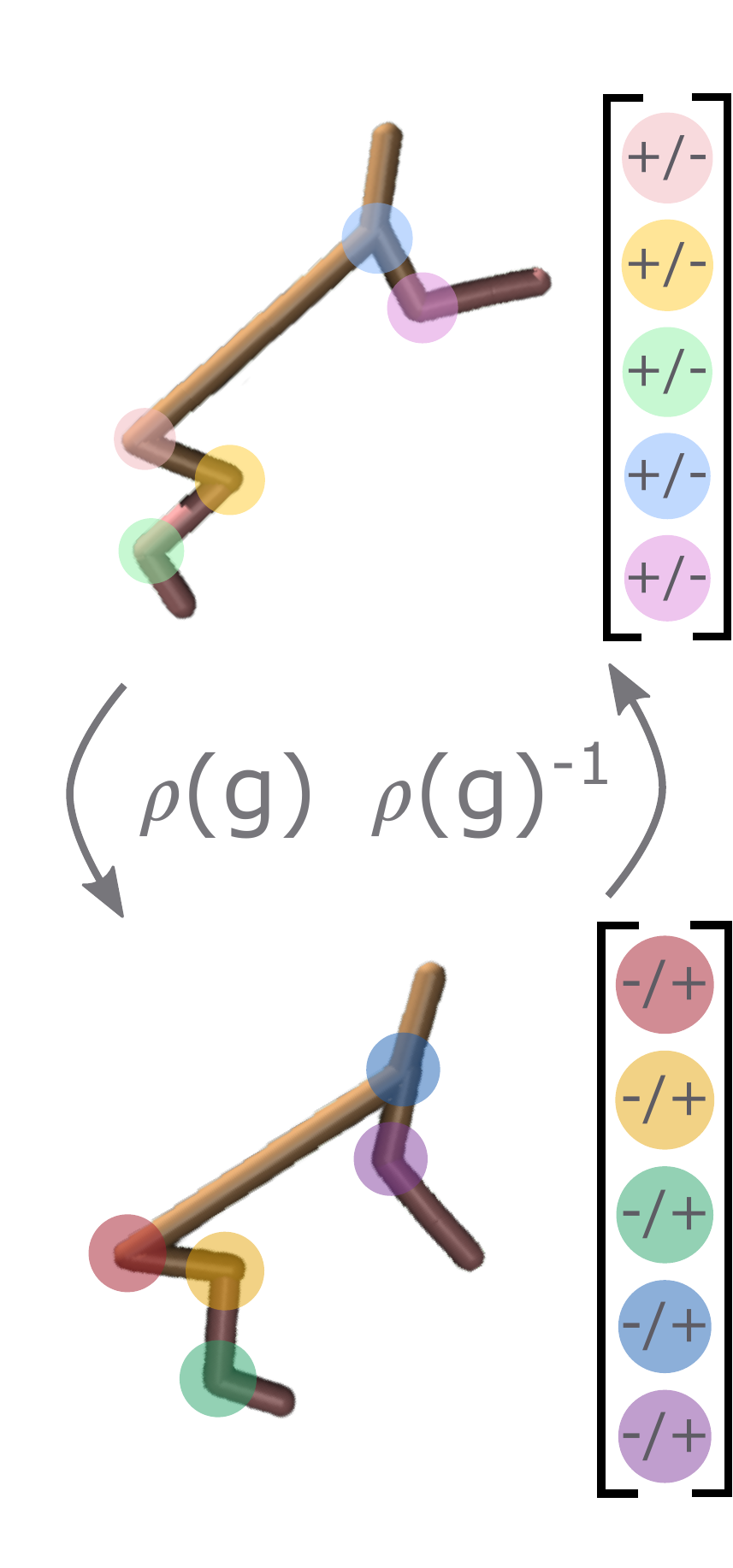}  
  \caption{HalfCheetah In-Out}
  \label{fig:misspec-sym}
\end{subfigure}
\caption{Example illustrations of symmetries and representations from the Mujoco environments. \textbf{Left:} left-right symmetry in the \emph{Walker2d} environment, \textbf{center:} front-back symmetry in the \emph{Swimmer} environment, and \textbf{right:} In-out similarity in the \emph{HalfCheetah} environment}
\label{fig:dynamics}
\end{figure}

We can make use of this information about what kinds of objects populate the state and action spaces to encode approximate symmetries of the RL environments. As shown in \citet{van2020mdp}, exploiting symmetries in MDPs by using equivariant networks can yield substantial improvements in data efficiency. 
But symmetries are brittle, and minor effects like rewards for moving in one direction, gravity, or even perturbations like wind, a minor tilt angle in CartPole, or other environment imperfections can break otherwise perfectly good symmetries. As shown in \autoref{table:symmetries}, broadening the scope to approximate symmetries allows for leveraging a lot more structure in the data which we can exploit with RPP. 
While Walker2d, Swimmer, Ant, and Humanoid have exact left/right reflection symmetries, Hopper, HalfCheetah, and Swimmer have approximate front/back reflection symmetries. Ant and Humanoid have an even more diverse set, with the $\mathrm{D}_4$ dihedral symmetry by reflecting and cyclicly permuting the legs of the ant, as well as continuous rotations of the Ant and Humanoid within the environment which can be broken by external forces or rewards. 
Identifying this structure in the data, we are able to use the generality of EMLP to construct an equivariant model for this data, and then turn it into a soft prior using RPP.

\begin{table}[h]
\centering
    \begin{tabular}{ccccccc}
        Symmetries & Walker2d & Hopper &  HalfCheetah& Swimmer &  Ant & Humanoid\\
         \midrule
         Exact & $\mathbb{Z}_2$& \ding{55}& \ding{55}& $\mathbb{Z}_2$&$\mathbb{Z}_2$ & $\mathbb{Z}_2$\\
         Approximate  & $\mathbb{Z}_2$& $\mathbb{Z}_2$& $\mathbb{Z}_2$& $\mathbb{Z}_2\times \mathbb{Z}_2$&$\mathrm{D}_4\times \mathrm{O}(2)$ & $\mathbb{Z}_2\times \mathrm{O}(2)$\\
         \midrule
         This work & $\mathbb{Z}_2$& $\mathbb{Z}_2$& $\mathbb{Z}_2$& $\mathbb{Z}_2\times \mathbb{Z}_2$&$\mathbb{Z}_4$ & $\mathrm{SO}(2)$\\
         
    \end{tabular}
    \vspace{2mm}
    \caption{Exact and approximate symmetries of Mujoco locomotion environments of which we use the subgroups in the bottom row, see \autoref{app:representations} for the detailed action and state representations.}
    \label{table:symmetries}
\end{table}

\begin{figure}
\begin{tabular}{ccc}
  \hspace*{-.33cm}\includegraphics[width=0.3\linewidth]{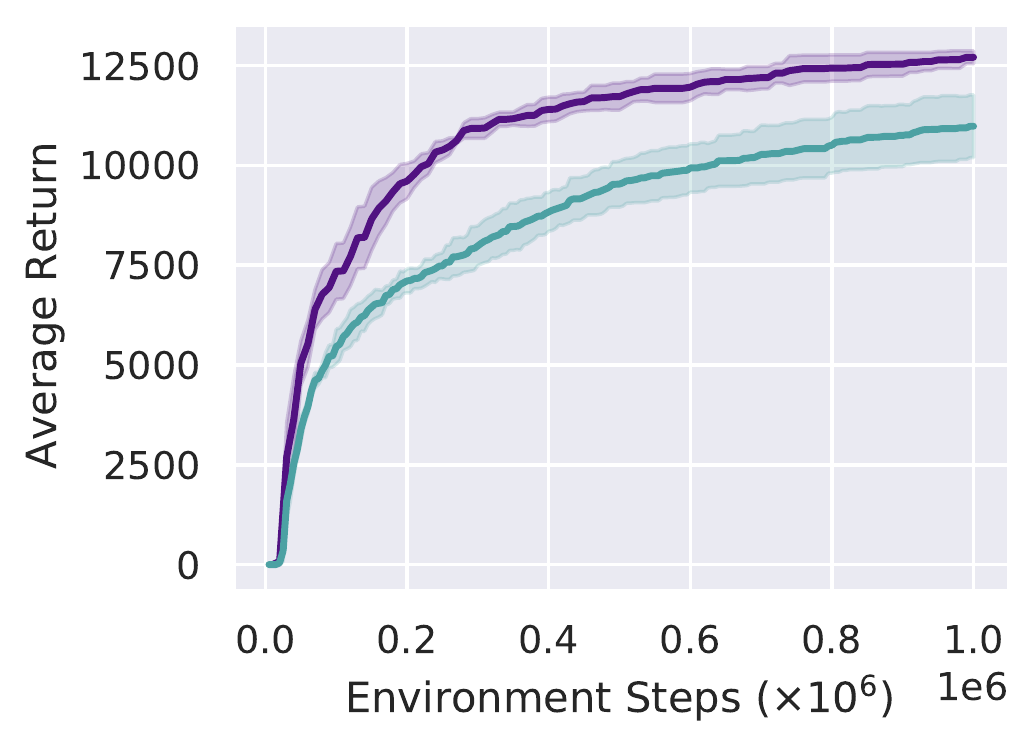} & 
  \hspace*{-.33cm}  \includegraphics[width=0.3\linewidth]{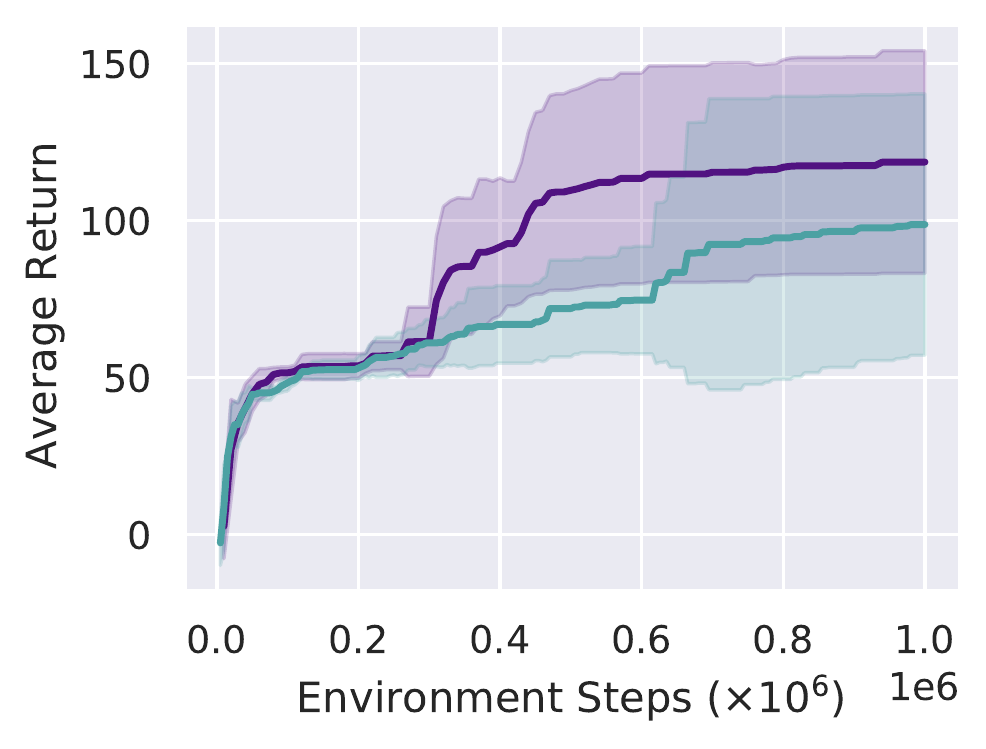}&\hspace*{-.33cm}\includegraphics[width=0.33\linewidth]{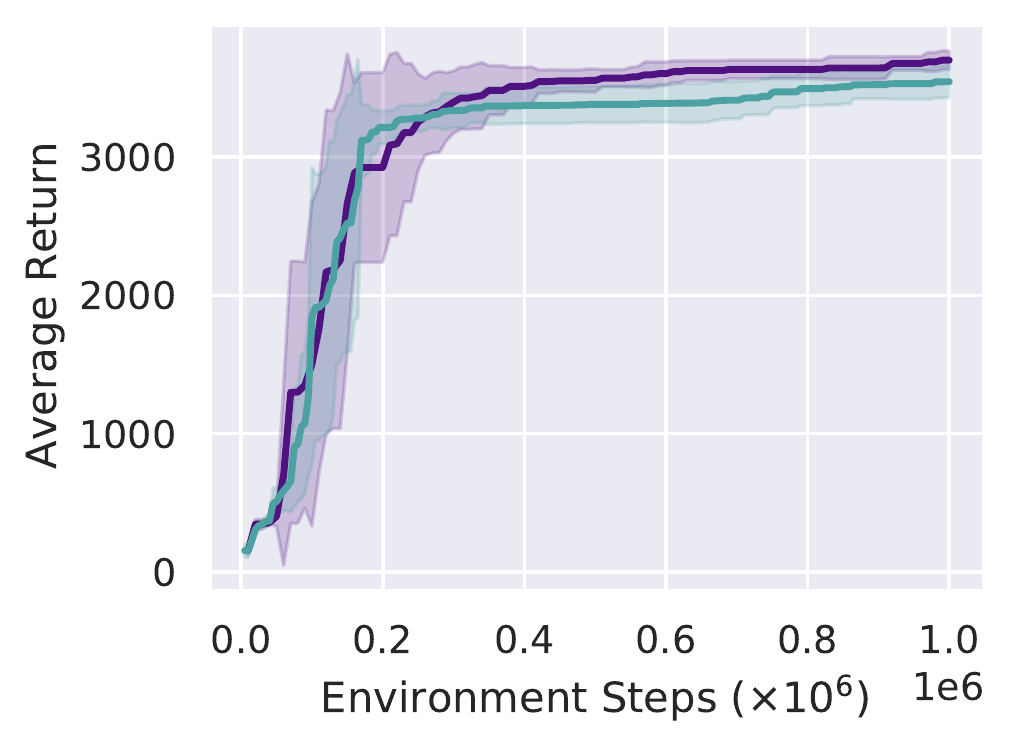} \\
HalfCheetah-v2 & Swimmer-v2 & Hopper-v2 \\[6pt]
 \hspace*{-.33cm}\includegraphics[width=0.33\linewidth]{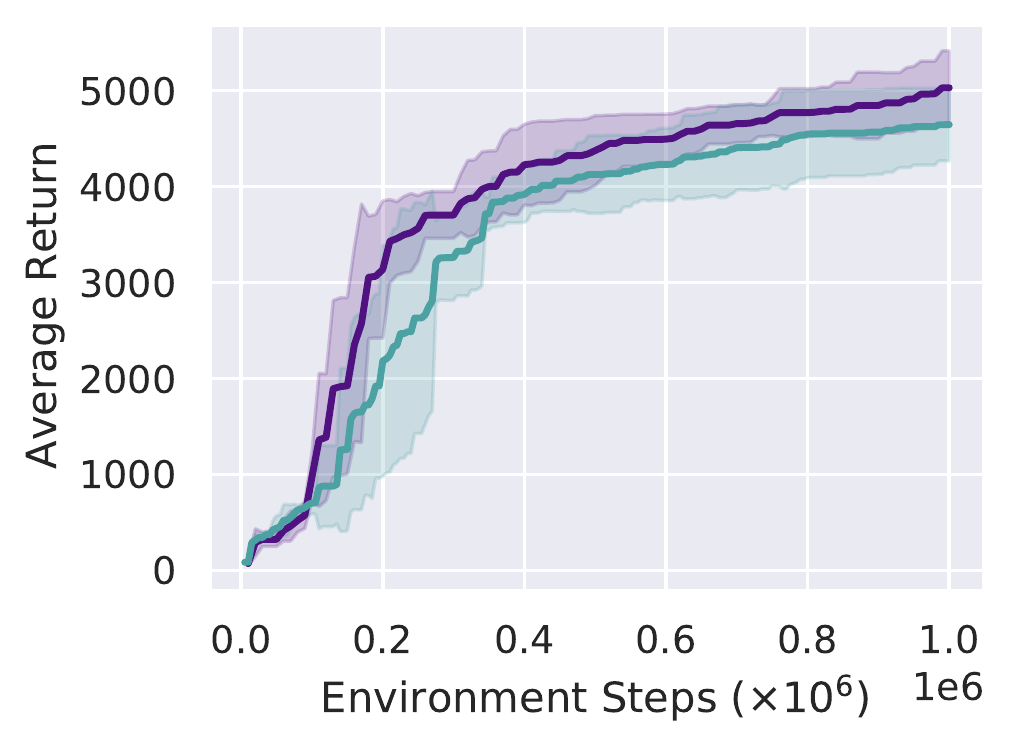} &   \hspace*{-.33cm}\includegraphics[width=0.33\linewidth]{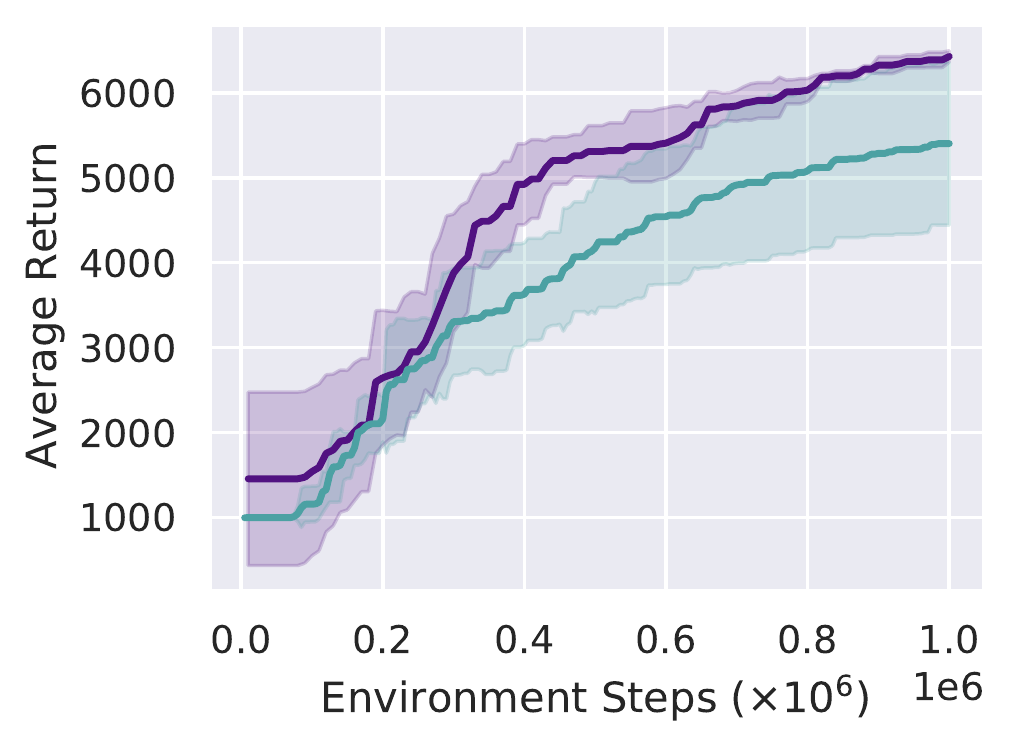}&
 \hspace*{-.33cm}\includegraphics[width=0.3\linewidth]{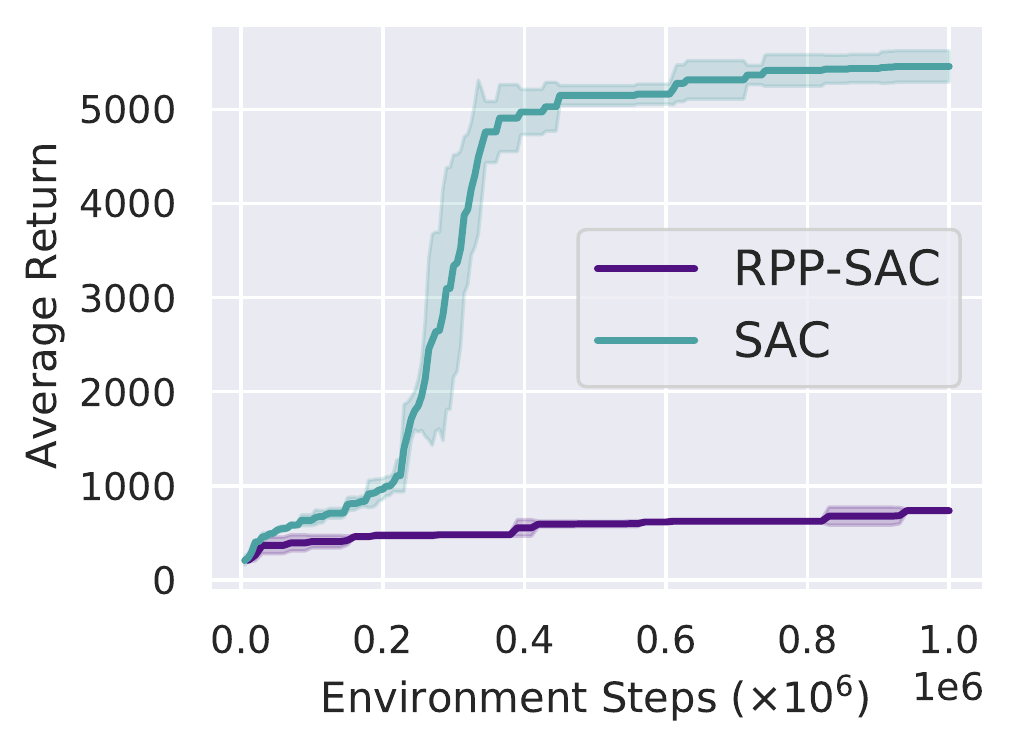}\\
Walker2d-v2& Ant-v2 & Humanoid-v2 \\[6pt]
\end{tabular}
\caption{Average reward curve of RPP-SAC and SAC trained on Mujoco locomotion environments (max average reward attained at each step). Mean and one standard deviation taken over $4$ trials shown in the shaded region. Incorporating approximate symmetries in the environments improves the efficiency of the model free RL agents.}
\label{fig:model-free-results}
\end{figure}

\subsection{Approximate Symmetries in Model Free Reinforcement Learning}\label{sec:mujoco-results}

We evaluate RPPs on the standard suite of Mujoco continuous control tasks in the context of model-free reinforcement learning. With the appropriately specified action and state representations detailed in \autoref{app:representations}, we construct RPP-EMLPs which we use as a drop-in replacement for both the policy and Q-function in the Soft Actor Critic (SAC) algorithm \citep{haarnoja2018soft}, using the same number of layers and channels. In contrast with \citet{van2020mdp} where equivariance is used just for policies, we find that using RPP-EMLP for the policy function alone is not very helpful with Actor Critic (see \autoref{fig:model-free-results}). With the exception of the Humanoid-v2 environment where the RPP-EMLP destabilizes SAC, we find that incorporating the exact and approximate equivariance with RPP yields consistent improvements in the data efficiency of the RL agent as shown in \autoref{fig:model-free-results}.

\subsection{Better Transition Models for Model Based Reinforcement Learning}
\begin{table}[h]
    \centering
    \begin{tabular}{ccccccc}
        \multicolumn{1}{c}{}&
        \multicolumn{2}{c}{Swimmer-v2}&
        \multicolumn{2}{c}{Hopper-v2}&
        \multicolumn{2}{c}{Ant-v2}\\
        \toprule
         Rollout & MLP & \bf{RPP} & MLP & \bf{RPP} & \bf{MLP} & RPP\\
         \midrule
        $10$ Steps & $0.51\pm0.02$  &  $\bf{0.40}\pm \bf{0.04}$  &  $1.1\pm0.1$  &  $\bf{0.9}\pm\bf{ 0.1}$  &  $\bf{4.2}\pm\bf{0.1}$  &  $5.2\pm 0.3$ \\
        $30$ Steps & $1.6\pm0.2$ & $\bf{1.26}\pm\bf{0.14}$ &  $3.8\pm0.3$ & $\bf{3.1}\pm\bf{ 0.5}$ & $\bf{11.3}\pm\bf{0.2}$ & $13.9\pm 0.7$ \\
        $100$ Steps & $3.9 \pm 1.0$ & $\bf{2.75}\pm\bf{0.31}$ & $9.8\pm0.5$ & $\bf{7.0}\pm\bf{ 0.7}$ & $\bf{16.0}\pm\bf{0.3}$ & $20.0\pm1.1$ \\
        \midrule
        Equiv Err & 46\%&19\%&98\%&32\%&36\%&31\% \\
    \end{tabular}
    \vspace{2mm}
    \caption{Transition model rollout relative error in percent \% averaged over $10$, $30$, and $100$ step rollouts  (geometric mean over trajectory). Errorbars are $1$ standard deviation taken over $3$ random seeds. Equivariance error is computed from as the geometric mean averaged over the $100$ step rollout.
    }
    \label{table:rollouts}
\end{table}
We also investigate whether the equivariance prior of RPP can improve the quality of the predictions for transition models in the context of model based RL. To evaluate this in a way decoupled from the complex interactions between policy, model, and value function in MBRL, we instead construct a static dataset of $50,000$ state transitions sampled uniformly from the replay buffer of a trained SAC agent. Since the trajectories in the replay buffer come from different times, they capture the varied dynamics MBRL transition models often encounter during training.

State of the art model based approaches on Mujoco tend to use an ensemble of small MLPs that predict the state transitions \citep{chua2018deep,wang2019exploring,janner2019trust,amos2020model}, without exploiting any structure of the state space.
We evaluate test rollout predictions via the relative error of the state over different length horizons for the RPP model against an MLP, the method of choice. As shown in \autoref{table:rollouts}, RPP transition models outperform MLPs on the Swimmer and Hopper environments, especially for long rollouts showing promise for use in MBRL. On these environments, RPP learns a smaller but non-negligible equivariance error that still enables it to fit the data.

\section{Limitations}\label{app: limitations}

Using RPP-EMLP for the state and action spaces of the Mujoco environments required identifying the meaning of each of the components in terms of whether they are scalars, velocity vectors, joint angles, or orientation quaternions, and also which part of the robot they correspond to. This can be an error-prone process. While RPPs are fairly robust to such mistakes, the need to identify components makes using RPP more challenging than standard MLP. Additionally, due to the bilinear layers within EMLP, the Lipschitz constant of the network is unbounded which can lead to training instabilities when the inputs are not well normalized. We hypothesize these factors may contribute to the training instability we experienced using RPP-EMLP on Humanoid-v2.

\section{Conclusion}
In this work we have presented a method for converting restriction priors such as equivariance constraints into flexible models that have a bias towards structure but are not constrained by it. Given uncertainty about the nature of the data, RPPs are a safe choice. These RPP models are able to perform as well as the equivariant models when exact symmetries are present, and as well as unstructured MLPs when the specified symmetry is absent, and better than both for approximate symmetries. We have shown that encoding approximate symmetries can be a powerful technique for improving performance in a variety of settings, particularly for 
the messy and complex state and action spaces in reinforcement learning.

We hope that RPP enables designing more expressive priors for neural networks that capture the kinds of high level assumptions that machine learning researchers actually hold when developing models, rather than low level characteristics about the parameters that are hard to interpret. Building better techniques for enforcing high level properties helps lower the cost of incorporating prior knowledge, and better accommodate the complexities of data, even if they don't match our expectations.  

\textbf{Acknowledgements}
We thank Samuel Stanton for useful discussion and feedback. This research was supported by  an  Amazon  Research  Award, NSF  I-DISRE  193471,  NIH R01DA048764-01A1,  NSF IIS-1910266,  and NSF 1922658NRT-HDR: FUTURE Foundations,  Translation,  and Responsibility for Data Science.

\bibliographystyle{plainnat}
\bibliography{refs}

\begin{thebibliography}{61}
\providecommand{\natexlab}[1]{#1}
\providecommand{\url}[1]{\texttt{#1}}
\expandafter\ifx\csname urlstyle\endcsname\relax
  \providecommand{\doi}[1]{doi: #1}\else
  \providecommand{\doi}{doi: \begingroup \urlstyle{rm}\Url}\fi

\bibitem[Abdolhosseini et~al.(2019)Abdolhosseini, Ling, Xie, Peng, and van~de
  Panne]{abdolhosseini2019learning}
Farzad Abdolhosseini, Hung~Yu Ling, Zhaoming Xie, Xue~Bin Peng, and Michiel
  van~de Panne.
\newblock On learning symmetric locomotion.
\newblock In \emph{Motion, Interaction and Games}, pages 1--10. 2019.

\bibitem[Amos et~al.(2020)Amos, Stanton, Yarats, and Wilson]{amos2020model}
Brandon Amos, Samuel Stanton, Denis Yarats, and Andrew~Gordon Wilson.
\newblock On the model-based stochastic value gradient for continuous
  reinforcement learning.
\newblock \emph{arXiv preprint arXiv:2008.12775}, 2020.

\bibitem[Anderson et~al.(2019)Anderson, Hy, and Kondor]{anderson2019cormorant}
Brandon Anderson, Truong~Son Hy, and Risi Kondor.
\newblock Cormorant: Covariant molecular neural networks.
\newblock In \emph{Advances in Neural Information Processing Systems}, pages
  14510--14519, 2019.

\bibitem[Andrychowicz et~al.(2020)Andrychowicz, Raichuk, Sta{\'n}czyk, Orsini,
  Girgin, Marinier, Hussenot, Geist, Pietquin, Michalski,
  et~al.]{andrychowicz2020matters}
Marcin Andrychowicz, Anton Raichuk, Piotr Sta{\'n}czyk, Manu Orsini, Sertan
  Girgin, Raphael Marinier, L{\'e}onard Hussenot, Matthieu Geist, Olivier
  Pietquin, Marcin Michalski, et~al.
\newblock What matters in on-policy reinforcement learning? a large-scale
  empirical study.
\newblock \emph{arXiv preprint arXiv:2006.05990}, 2020.

\bibitem[Benton et~al.(2020)Benton, Finzi, Izmailov, and
  Wilson]{benton2020learning}
Gregory Benton, Marc Finzi, Pavel Izmailov, and Andrew~Gordon Wilson.
\newblock Learning invariances in neural networks.
\newblock \emph{arXiv preprint arXiv:2010.11882}, 2020.

\bibitem[Bogatskiy et~al.(2020)Bogatskiy, Anderson, Offermann, Roussi, Miller,
  and Kondor]{bogatskiy2020lorentz}
Alexander Bogatskiy, Brandon Anderson, Jan Offermann, Marwah Roussi, David
  Miller, and Risi Kondor.
\newblock Lorentz group equivariant neural network for particle physics.
\newblock In \emph{International Conference on Machine Learning}, pages
  992--1002. PMLR, 2020.

\bibitem[Brockman et~al.(2016)Brockman, Cheung, Pettersson, Schneider,
  Schulman, Tang, and Zaremba]{brockman2016openai}
Greg Brockman, Vicki Cheung, Ludwig Pettersson, Jonas Schneider, John Schulman,
  Jie Tang, and Wojciech Zaremba.
\newblock Openai gym.
\newblock \emph{arXiv preprint arXiv:1606.01540}, 2016.

\bibitem[Chen et~al.(2018)Chen, Rubanova, Bettencourt, and
  Duvenaud]{chen2018neural}
Ricky~TQ Chen, Yulia Rubanova, Jesse Bettencourt, and David~K Duvenaud.
\newblock Neural ordinary differential equations.
\newblock In \emph{Advances in neural information processing systems}, pages
  6571--6583, 2018.

\bibitem[Chua et~al.(2018)Chua, Calandra, McAllister, and Levine]{chua2018deep}
Kurtland Chua, Roberto Calandra, Rowan McAllister, and Sergey Levine.
\newblock Deep reinforcement learning in a handful of trials using
  probabilistic dynamics models.
\newblock \emph{arXiv preprint arXiv:1805.12114}, 2018.

\bibitem[Cohen and Welling(2016)]{cohen2016group}
Taco Cohen and Max Welling.
\newblock Group equivariant convolutional networks.
\newblock In \emph{International conference on machine learning}, pages
  2990--2999. PMLR, 2016.

\bibitem[Dai et~al.(2021)Dai, Liu, Le, and Tan]{dai2021coatnet}
Zihang Dai, Hanxiao Liu, Quoc~V Le, and Mingxing Tan.
\newblock Coatnet: Marrying convolution and attention for all data sizes.
\newblock \emph{arXiv preprint arXiv:2106.04803}, 2021.

\bibitem[d'Ascoli et~al.(2021)d'Ascoli, Touvron, Leavitt, Morcos, Biroli, and
  Sagun]{d2021convit}
St{\'e}phane d'Ascoli, Hugo Touvron, Matthew Leavitt, Ari Morcos, Giulio
  Biroli, and Levent Sagun.
\newblock Convit: Improving vision transformers with soft convolutional
  inductive biases.
\newblock \emph{arXiv preprint arXiv:2103.10697}, 2021.

\bibitem[Dosovitskiy et~al.(2020)Dosovitskiy, Beyer, Kolesnikov, Weissenborn,
  Zhai, Unterthiner, Dehghani, Minderer, Heigold, Gelly,
  et~al.]{dosovitskiy2020image}
Alexey Dosovitskiy, Lucas Beyer, Alexander Kolesnikov, Dirk Weissenborn,
  Xiaohua Zhai, Thomas Unterthiner, Mostafa Dehghani, Matthias Minderer, Georg
  Heigold, Sylvain Gelly, et~al.
\newblock An image is worth 16x16 words: Transformers for image recognition at
  scale.
\newblock \emph{arXiv preprint arXiv:2010.11929}, 2020.

\bibitem[Dua and Graff(2017)]{Dua:2019}
Dheeru Dua and Casey Graff.
\newblock {UCI} machine learning repository, 2017.
\newblock URL \url{http://archive.ics.uci.edu/ml}.

\bibitem[Elesedy and Zaidi(2021)]{elesedy2021provably}
Bryn Elesedy and Sheheryar Zaidi.
\newblock Provably strict generalisation benefit for equivariant models.
\newblock \emph{arXiv preprint arXiv:2102.10333}, 2021.

\bibitem[Finzi et~al.(2020)Finzi, Stanton, Izmailov, and
  Wilson]{finzi2020generalizing}
Marc Finzi, Samuel Stanton, Pavel Izmailov, and Andrew~Gordon Wilson.
\newblock Generalizing convolutional neural networks for equivariance to lie
  groups on arbitrary continuous data.
\newblock In \emph{International Conference on Machine Learning}, pages
  3165--3176. PMLR, 2020.

\bibitem[Finzi et~al.(2021)Finzi, Welling, and Wilson]{finzi2021practical}
Marc Finzi, Max Welling, and Andrew~Gordon Wilson.
\newblock A practical method for constructing equivariant multilayer
  perceptrons for arbitrary matrix groups.
\newblock \emph{arXiv preprint arXiv:2104.09459}, 2021.

\bibitem[Fuchs et~al.(2020)Fuchs, Worrall, Fischer, and Welling]{fuchs2020se}
Fabian~B Fuchs, Daniel~E Worrall, Volker Fischer, and Max Welling.
\newblock Se (3)-transformers: 3d roto-translation equivariant attention
  networks.
\newblock \emph{arXiv preprint arXiv:2006.10503}, 2020.

\bibitem[Greydanus et~al.(2019)Greydanus, Dzamba, and
  Yosinski]{greydanus2019hamiltonian}
Samuel Greydanus, Misko Dzamba, and Jason Yosinski.
\newblock Hamiltonian neural networks.
\newblock In \emph{Advances in Neural Information Processing Systems}, pages
  15379--15389, 2019.

\bibitem[Haarnoja et~al.(2018{\natexlab{a}})Haarnoja, Zhou, Abbeel, and
  Levine]{haarnoja2018soft}
Tuomas Haarnoja, Aurick Zhou, Pieter Abbeel, and Sergey Levine.
\newblock Soft actor-critic: Off-policy maximum entropy deep reinforcement
  learning with a stochastic actor.
\newblock In \emph{International Conference on Machine Learning}, pages
  1861--1870. PMLR, 2018{\natexlab{a}}.

\bibitem[Haarnoja et~al.(2018{\natexlab{b}})Haarnoja, Zhou, Abbeel, and
  Levine]{haarnoja2018soft2}
Tuomas Haarnoja, Aurick Zhou, Pieter Abbeel, and Sergey Levine.
\newblock Soft actor-critic: Off-policy maximum entropy deep reinforcement
  learning with a stochastic actor.
\newblock In \emph{International Conference on Machine Learning}, pages
  1861--1870. PMLR, 2018{\natexlab{b}}.

\bibitem[Haarnoja et~al.(2018{\natexlab{c}})Haarnoja, Zhou, Hartikainen,
  Tucker, Ha, Tan, Kumar, Zhu, Gupta, Abbeel, et~al.]{haarnoja2018soft1}
Tuomas Haarnoja, Aurick Zhou, Kristian Hartikainen, George Tucker, Sehoon Ha,
  Jie Tan, Vikash Kumar, Henry Zhu, Abhishek Gupta, Pieter Abbeel, et~al.
\newblock Soft actor-critic algorithms and applications.
\newblock \emph{arXiv preprint arXiv:1812.05905}, 2018{\natexlab{c}}.

\bibitem[He et~al.(2016{\natexlab{a}})He, Zhang, Ren, and Sun]{he2016deep}
Kaiming He, Xiangyu Zhang, Shaoqing Ren, and Jian Sun.
\newblock Deep residual learning for image recognition.
\newblock In \emph{Proceedings of the IEEE conference on computer vision and
  pattern recognition}, pages 770--778, 2016{\natexlab{a}}.

\bibitem[He et~al.(2016{\natexlab{b}})He, Zhang, Ren, and Sun]{he2016identity}
Kaiming He, Xiangyu Zhang, Shaoqing Ren, and Jian Sun.
\newblock Identity mappings in deep residual networks.
\newblock In \emph{European conference on computer vision}, pages 630--645.
  Springer, 2016{\natexlab{b}}.

\bibitem[Janner et~al.(2019)Janner, Fu, Zhang, and Levine]{janner2019trust}
Michael Janner, Justin Fu, Marvin Zhang, and Sergey Levine.
\newblock When to trust your model: Model-based policy optimization.
\newblock \emph{arXiv preprint arXiv:1906.08253}, 2019.

\bibitem[Jiang et~al.(2018)Jiang, Dun, Huang, and Lu]{jiang2018graph}
Jiechuan Jiang, Chen Dun, Tiejun Huang, and Zongqing Lu.
\newblock Graph convolutional reinforcement learning.
\newblock \emph{arXiv preprint arXiv:1810.09202}, 2018.

\bibitem[Johannink et~al.(2019)Johannink, Bahl, Nair, Luo, Kumar, Loskyll,
  Ojea, Solowjow, and Levine]{johannink2019residual}
Tobias Johannink, Shikhar Bahl, Ashvin Nair, Jianlan Luo, Avinash Kumar,
  Matthias Loskyll, Juan~Aparicio Ojea, Eugen Solowjow, and Sergey Levine.
\newblock Residual reinforcement learning for robot control.
\newblock In \emph{2019 International Conference on Robotics and Automation
  (ICRA)}, pages 6023--6029. IEEE, 2019.

\bibitem[Kashinath et~al.(2021)Kashinath, Mustafa, Albert, Wu, Jiang,
  Esmaeilzadeh, Azizzadenesheli, Wang, Chattopadhyay, Singh,
  et~al.]{kashinath2021physics}
K~Kashinath, M~Mustafa, A~Albert, JL~Wu, C~Jiang, S~Esmaeilzadeh,
  K~Azizzadenesheli, R~Wang, A~Chattopadhyay, A~Singh, et~al.
\newblock Physics-informed machine learning: case studies for weather and
  climate modelling.
\newblock \emph{Philosophical Transactions of the Royal Society A},
  379\penalty0 (2194):\penalty0 20200093, 2021.

\bibitem[Kingma and Ba(2014)]{kingma2014adam}
Diederik~P Kingma and Jimmy Ba.
\newblock Adam: A method for stochastic optimization.
\newblock \emph{arXiv preprint arXiv:1412.6980}, 2014.

\bibitem[Kostrikov et~al.(2020)Kostrikov, Yarats, and
  Fergus]{kostrikov2020image}
Ilya Kostrikov, Denis Yarats, and Rob Fergus.
\newblock Image augmentation is all you need: Regularizing deep reinforcement
  learning from pixels.
\newblock \emph{arXiv preprint arXiv:2004.13649}, 2020.

\bibitem[Krizhevsky et~al.(2009)Krizhevsky, Hinton,
  et~al.]{krizhevsky2009learning}
Alex Krizhevsky, Geoffrey Hinton, et~al.
\newblock Learning multiple layers of features from tiny images.
\newblock 2009.

\bibitem[LeCun et~al.(1989)LeCun, Boser, Denker, Henderson, Howard, Hubbard,
  and Jackel]{lecun1989backpropagation}
Yann LeCun, Bernhard Boser, John~S Denker, Donnie Henderson, Richard~E Howard,
  Wayne Hubbard, and Lawrence~D Jackel.
\newblock Backpropagation applied to handwritten zip code recognition.
\newblock \emph{Neural computation}, 1\penalty0 (4):\penalty0 541--551, 1989.

\bibitem[Lin et~al.(2020)Lin, Huang, Zimmer, Guan, Rojas, and
  Weng]{lin2020invariant}
Yijiong Lin, Jiancong Huang, Matthieu Zimmer, Yisheng Guan, Juan Rojas, and
  Paul Weng.
\newblock Invariant transform experience replay: Data augmentation for deep
  reinforcement learning.
\newblock \emph{IEEE Robotics and Automation Letters}, 5\penalty0 (4):\penalty0
  6615--6622, 2020.

\bibitem[Liu et~al.(2020)Liu, Yeh, and Schwing]{liu2020pic}
Iou-Jen Liu, Raymond~A Yeh, and Alexander~G Schwing.
\newblock Pic: permutation invariant critic for multi-agent deep reinforcement
  learning.
\newblock In \emph{Conference on Robot Learning}, pages 590--602. PMLR, 2020.

\bibitem[Liu et~al.(2018)Liu, Lehman, Molino, Such, Frank, Sergeev, and
  Yosinski]{liu2018intriguing}
Rosanne Liu, Joel Lehman, Piero Molino, Felipe~Petroski Such, Eric Frank, Alex
  Sergeev, and Jason Yosinski.
\newblock An intriguing failing of convolutional neural networks and the
  coordconv solution.
\newblock \emph{arXiv preprint arXiv:1807.03247}, 2018.

\bibitem[Liu et~al.(2021)Liu, Chen, Du, and Tegmark]{liu2021physics}
Ziming Liu, Yunyue Chen, Yuanqi Du, and Max Tegmark.
\newblock Physics-augmented learning: A new paradigm beyond physics-informed
  learning.
\newblock \emph{arXiv preprint arXiv:2109.13901}, 2021.

\bibitem[MacKay and Mac~Kay(2003)]{mackay2003information}
David~JC MacKay and David~JC Mac~Kay.
\newblock \emph{Information theory, inference and learning algorithms}.
\newblock Cambridge university press, 2003.

\bibitem[Maron et~al.(2018)Maron, Ben-Hamu, Shamir, and
  Lipman]{maron2018invariant}
Haggai Maron, Heli Ben-Hamu, Nadav Shamir, and Yaron Lipman.
\newblock Invariant and equivariant graph networks.
\newblock \emph{arXiv preprint arXiv:1812.09902}, 2018.

\bibitem[Maron et~al.(2020)Maron, Litany, Chechik, and
  Fetaya]{maron2020learning}
Haggai Maron, Or~Litany, Gal Chechik, and Ethan Fetaya.
\newblock On learning sets of symmetric elements.
\newblock \emph{arXiv preprint arXiv:2002.08599}, 2020.

\bibitem[Mavalankar(2020)]{mavalankar2020goal}
Aditi Mavalankar.
\newblock Goal-conditioned batch reinforcement learning for rotation invariant
  locomotion.
\newblock \emph{arXiv preprint arXiv:2004.08356}, 2020.

\bibitem[Miyato et~al.(2018)Miyato, Kataoka, Koyama, and
  Yoshida]{miyato2018spectral}
Takeru Miyato, Toshiki Kataoka, Masanori Koyama, and Yuichi Yoshida.
\newblock Spectral normalization for generative adversarial networks.
\newblock \emph{arXiv preprint arXiv:1802.05957}, 2018.

\bibitem[Neyshabur(2020)]{neyshabur2020towards}
Behnam Neyshabur.
\newblock Towards learning convolutions from scratch.
\newblock \emph{arXiv preprint arXiv:2007.13657}, 2020.

\bibitem[Neyshabur et~al.(2014)Neyshabur, Tomioka, and
  Srebro]{neyshabur2014search}
Behnam Neyshabur, Ryota Tomioka, and Nathan Srebro.
\newblock In search of the real inductive bias: On the role of implicit
  regularization in deep learning.
\newblock \emph{arXiv preprint arXiv:1412.6614}, 2014.

\bibitem[Ramachandran et~al.(2017)Ramachandran, Zoph, and
  Le]{ramachandran2017searching}
Prajit Ramachandran, Barret Zoph, and Quoc~V Le.
\newblock Searching for activation functions.
\newblock \emph{arXiv preprint arXiv:1710.05941}, 2017.

\bibitem[Ravindran and Barto(2004)]{ravindran2004approximate}
Balaraman Ravindran and Andrew~G Barto.
\newblock Approximate homomorphisms: A framework for non-exact minimization in
  markov decision processes.
\newblock 2004.

\bibitem[Rumelhart et~al.(1985)Rumelhart, Hinton, and
  Williams]{rumelhart1985learning}
David~E Rumelhart, Geoffrey~E Hinton, and Ronald~J Williams.
\newblock Learning internal representations by error propagation.
\newblock Technical report, California Univ San Diego La Jolla Inst for
  Cognitive Science, 1985.

\bibitem[Satorras et~al.(2021)Satorras, Hoogeboom, and Welling]{satorras2021n}
Victor~Garcia Satorras, Emiel Hoogeboom, and Max Welling.
\newblock E (n) equivariant graph neural networks.
\newblock \emph{arXiv preprint arXiv:2102.09844}, 2021.

\bibitem[Silver et~al.(2018)Silver, Allen, Tenenbaum, and
  Kaelbling]{silver2018residual}
Tom Silver, Kelsey Allen, Josh Tenenbaum, and Leslie Kaelbling.
\newblock Residual policy learning.
\newblock \emph{arXiv preprint arXiv:1812.06298}, 2018.

\bibitem[Sukhbaatar et~al.(2016)Sukhbaatar, Szlam, and
  Fergus]{sukhbaatar2016learning}
Sainbayar Sukhbaatar, Arthur Szlam, and Rob Fergus.
\newblock Learning multiagent communication with backpropagation.
\newblock \emph{arXiv preprint arXiv:1605.07736}, 2016.

\bibitem[Taylor et~al.(2008)Taylor, Precup, and Panagaden]{taylor2008bounding}
Jonathan Taylor, Doina Precup, and Prakash Panagaden.
\newblock Bounding performance loss in approximate mdp homomorphisms.
\newblock \emph{Advances in Neural Information Processing Systems},
  21:\penalty0 1649--1656, 2008.

\bibitem[van~der Pol et~al.(2020{\natexlab{a}})van~der Pol, Kipf, Oliehoek, and
  Welling]{van2020plannable}
Elise van~der Pol, Thomas Kipf, Frans~A Oliehoek, and Max Welling.
\newblock Plannable approximations to mdp homomorphisms: Equivariance under
  actions.
\newblock \emph{arXiv preprint arXiv:2002.11963}, 2020{\natexlab{a}}.

\bibitem[van~der Pol et~al.(2020{\natexlab{b}})van~der Pol, Worrall, van Hoof,
  Oliehoek, and Welling]{van2020mdp}
Elise van~der Pol, Daniel Worrall, Herke van Hoof, Frans Oliehoek, and Max
  Welling.
\newblock Mdp homomorphic networks: Group symmetries in reinforcement learning.
\newblock \emph{Advances in Neural Information Processing Systems}, 33,
  2020{\natexlab{b}}.

\bibitem[van~der Wilk et~al.(2018)van~der Wilk, Bauer, John, and
  Hensman]{van2018learning}
Mark van~der Wilk, Matthias Bauer, ST~John, and James Hensman.
\newblock Learning invariances using the marginal likelihood.
\newblock \emph{arXiv preprint arXiv:1808.05563}, 2018.

\bibitem[Wang et~al.(2020)Wang, Albooyeh, and Ravanbakhsh]{wang2020equivariant}
Renhao Wang, Marjan Albooyeh, and Siamak Ravanbakhsh.
\newblock Equivariant maps for hierarchical structures.
\newblock \emph{arXiv preprint arXiv:2006.03627}, 2020.

\bibitem[Wang and Ba(2019)]{wang2019exploring}
Tingwu Wang and Jimmy Ba.
\newblock Exploring model-based planning with policy networks.
\newblock \emph{arXiv preprint arXiv:1906.08649}, 2019.

\bibitem[Weiler and Cesa(2019)]{weiler2019general}
Maurice Weiler and Gabriele Cesa.
\newblock General $ e (2) $-equivariant steerable cnns.
\newblock \emph{arXiv preprint arXiv:1911.08251}, 2019.

\bibitem[Wilson and Izmailov(2020)]{wilson2020bayesian}
Andrew~Gordon Wilson and Pavel Izmailov.
\newblock Bayesian deep learning and a probabilistic perspective of
  generalization.
\newblock In \emph{Advances in Neural Information Processing Systems}, 2020.

\bibitem[Worrall et~al.(2017)Worrall, Garbin, Turmukhambetov, and
  Brostow]{worrall2017harmonic}
Daniel~E Worrall, Stephan~J Garbin, Daniyar Turmukhambetov, and Gabriel~J
  Brostow.
\newblock Harmonic networks: Deep translation and rotation equivariance.
\newblock In \emph{Proceedings of the IEEE Conference on Computer Vision and
  Pattern Recognition}, pages 5028--5037, 2017.

\bibitem[Xiao et~al.(2021)Xiao, Dollar, Singh, Mintun, Darrell, and
  Girshick]{xiao2021early}
Tete Xiao, Piotr Dollar, Mannat Singh, Eric Mintun, Trevor Darrell, and Ross
  Girshick.
\newblock Early convolutions help transformers see better.
\newblock In \emph{Thirty-Fifth Conference on Neural Information Processing
  Systems}, 2021.

\bibitem[Zaheer et~al.(2017)Zaheer, Kottur, Ravanbakhsh, Poczos, Salakhutdinov,
  and Smola]{zaheer2017deep}
Manzil Zaheer, Satwik Kottur, Siamak Ravanbakhsh, Barnabas Poczos, Russ~R
  Salakhutdinov, and Alexander~J Smola.
\newblock Deep sets.
\newblock In \emph{Advances in neural information processing systems}, pages
  3391--3401, 2017.

\bibitem[Zhou et~al.(2017)Zhou, Ye, Qiu, and Jiao]{zhou2017oriented}
Yanzhao Zhou, Qixiang Ye, Qiang Qiu, and Jianbin Jiao.
\newblock Oriented response networks.
\newblock In \emph{Proceedings of the IEEE Conference on Computer Vision and
  Pattern Recognition}, pages 519--528, 2017.

\end{thebibliography}

\newpage
\appendix

\section*{\centering Residual Pathway Priors for Soft Equivariance Constraints \\ Supplementary Material}

\section*{Appendix Outline}

In Section \ref{app: limitations} discuss potential for negative impact. 
In Section \ref{app:policy_only} we investigate the utility of using RPP-EMLP for the policy function only on the Mujoco tasks.
In Section \ref{app:experimental-details} we detail the datasets and experimental methodology used in the paper.
Finally in Sections \ref{app:representations} and \ref{app:spaces} we break down the components of the Mujoco environment state and action spaces, and the representations that we use for them.

\section{Potential Negative Impacts}
As one of our primary application areas is reinforcement learning, and specifically exploiting approximate symmetries in reinforcement learning, we must address the potential negative impacts of the deployment of RPPs in RL systems. In general model free RL algorithms tend to be brittle, and often policies and behavior learned in a simulated environment like Mujoco don't transfer easily to real world robots. This point is acknowledged by most RL researchers, and a large effort is being made to improve the situation. Applying neural networks to the control of real robots can be dangerous if the functions are important or failure can cause injury to the robot or humans. We believe that RL will ultimately be impactful for robot control, however practitioners need to be responsible and exercise caution.

\section{Benefit of Equivariant Value Functions}\label{app:policy_only}
In principle both the policy and the value or critic function can benefit from equivariance. However, the policy learns from the value function in the policy update which is approximately equivalent to minimizing the KL divergence
\begin{equation*}
    \mathbb{E}_{s\sim \mathcal{D}}[ \mathrm{KL}(\pi_\phi(\cdot|s)|\exp(Q_\theta(\cdot,s))/Z_\theta(s))]
\end{equation*}
as derived in \citet{haarnoja2018soft2}. If the value function $Q$ is a standard MLP yielding a non equivariant distribution and the policy function $\pi$ is an RPP that merely has a bias towards equivariance, then the RPP policy will learn to fit the non equivariant parts of $Q$ as if it were a ground truth dataset that is not equivariant. This likely explains why we find in practice that using an RPP for the value function has a stronger impact on performance as shown in \autoref{fig:model-free-results}.

\begin{figure}
\begin{tabular}{ccc}
  \hspace*{-.33cm}\includegraphics[width=0.34\linewidth]{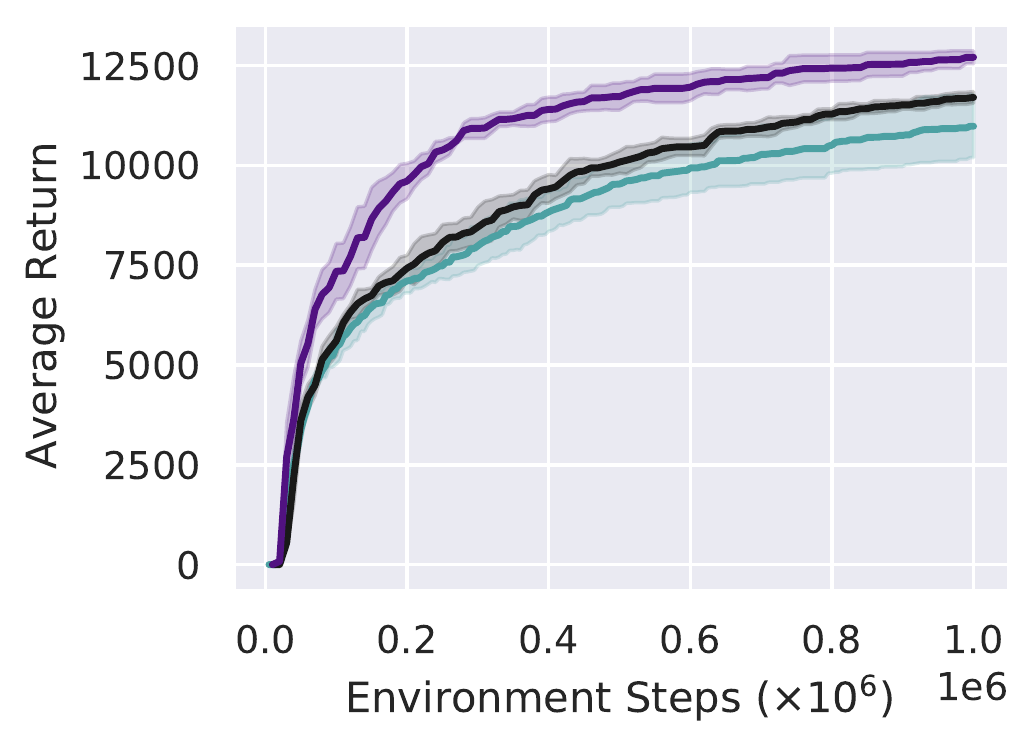} & 
  \hspace*{-.33cm}  \includegraphics[width=0.32\linewidth]{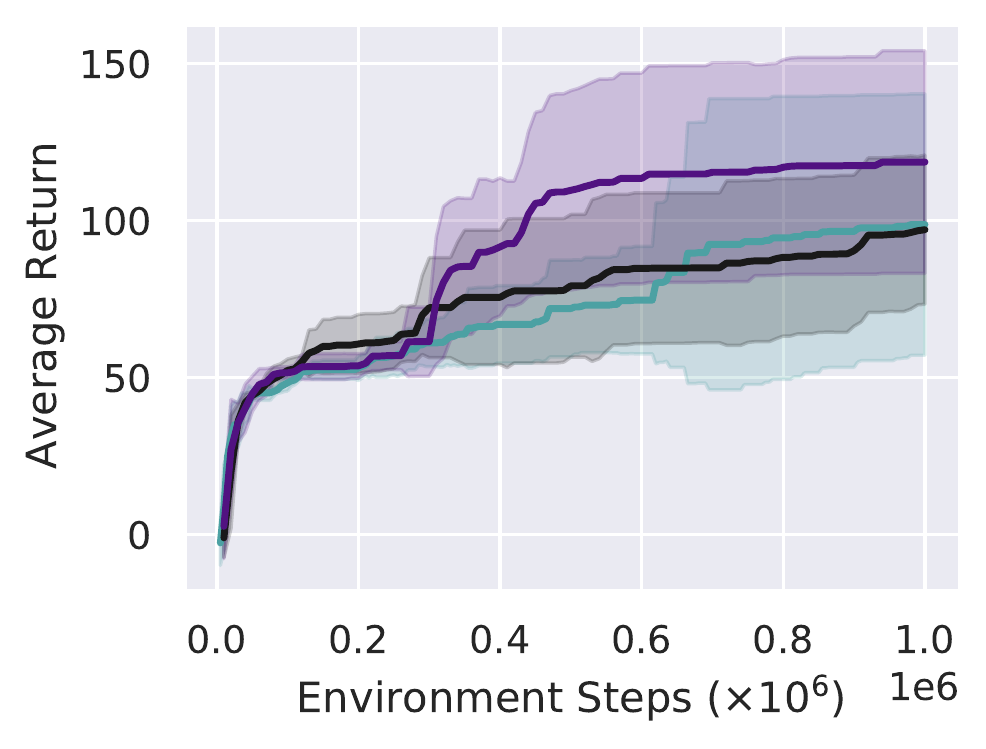}&\hspace*{-.33cm}\includegraphics[width=0.33\linewidth]{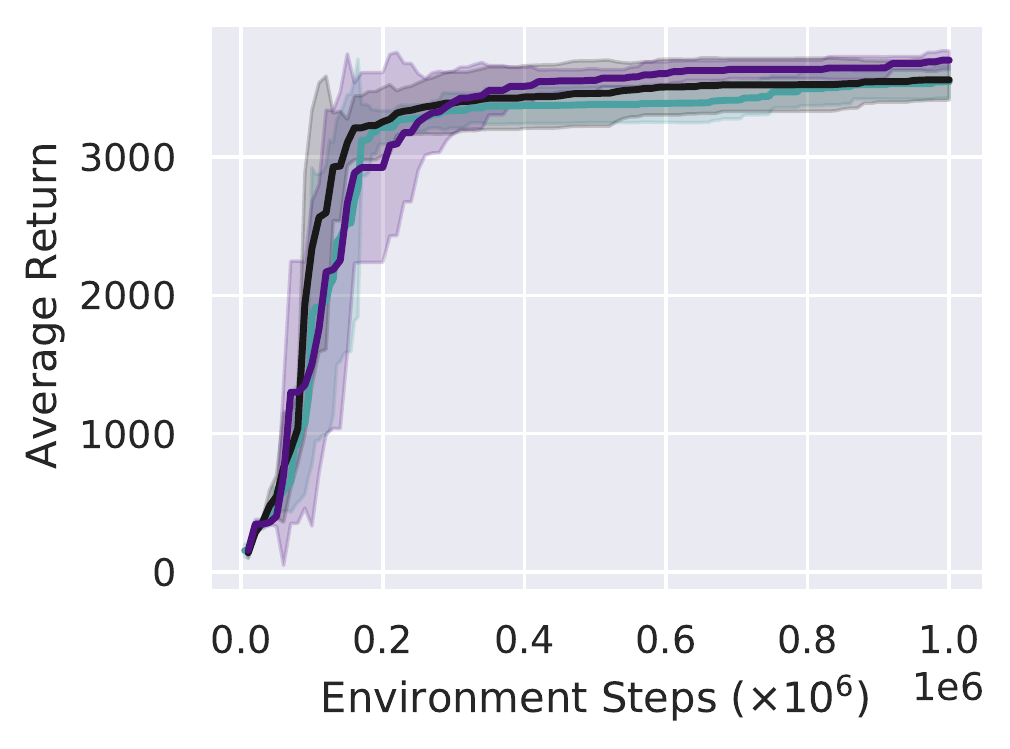} \\
HalfCheetah-v2 & Swimmer-v2 & Hopper-v2 \\[6pt]
 \hspace*{-.33cm}\includegraphics[width=0.33\linewidth]{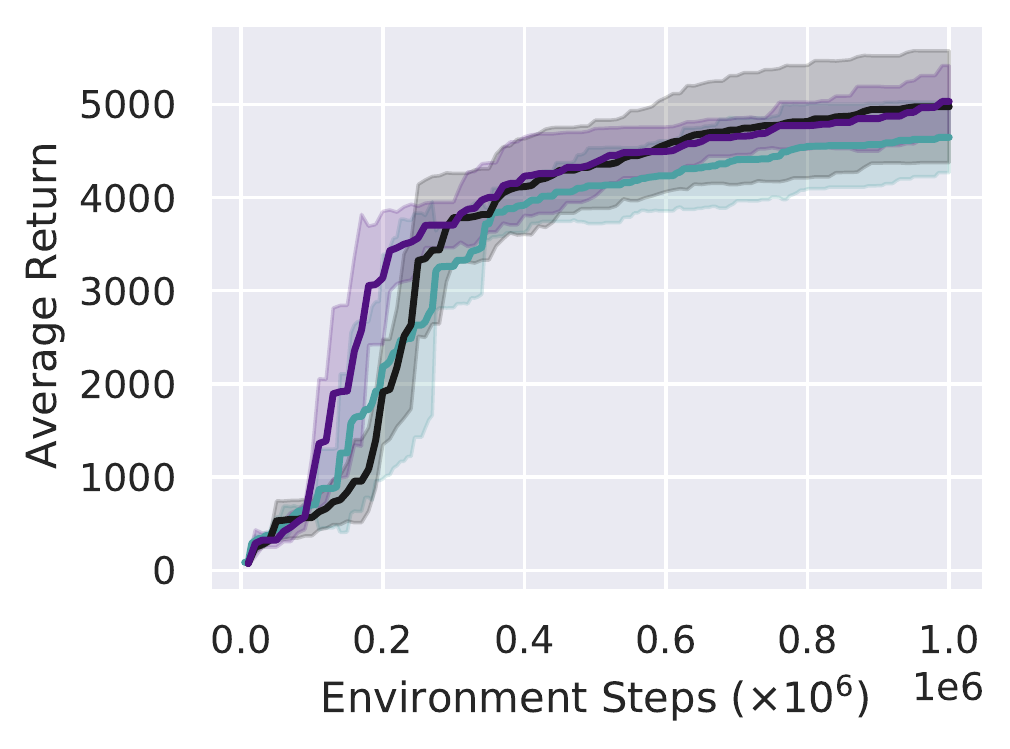} &   \hspace*{-.33cm}\includegraphics[width=0.33\linewidth]{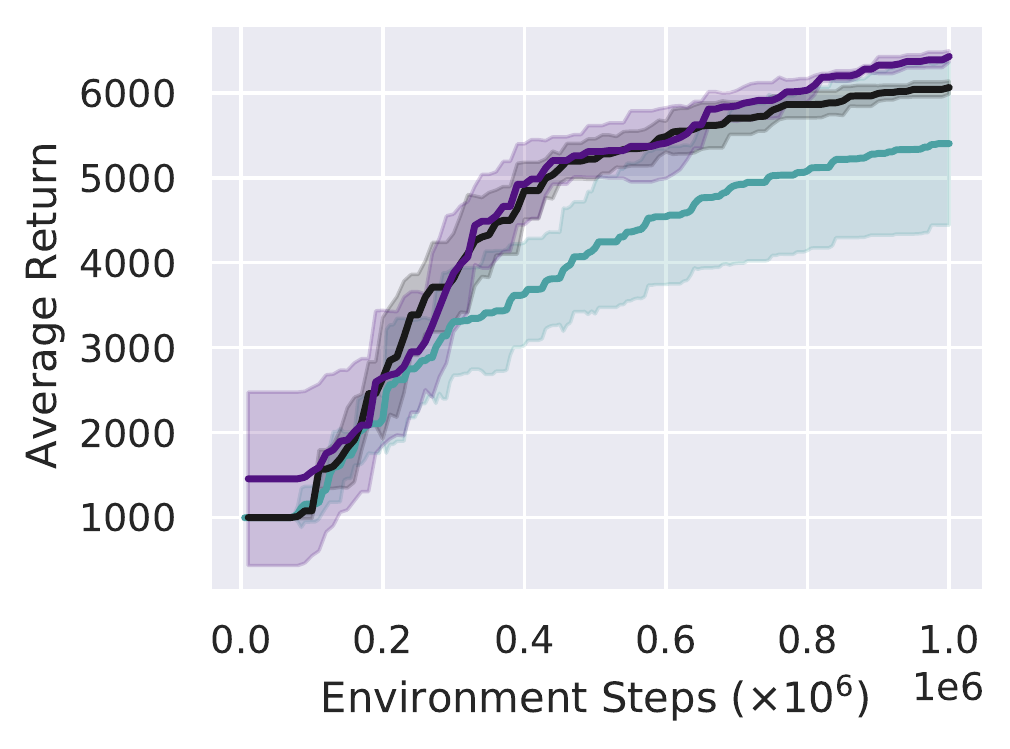}&
 \hspace*{-.33cm}\includegraphics[width=0.33\linewidth]{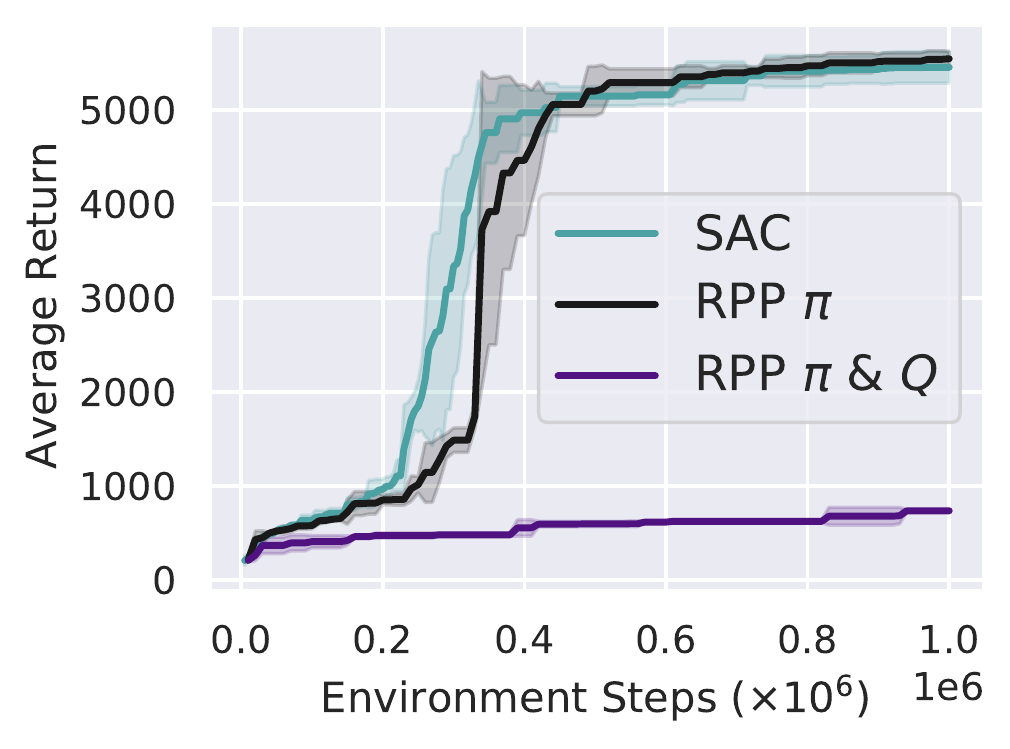}\\
Walker2d-v2& Ant-v2 & Humanoid-v2 \\[6pt]
\end{tabular}
\caption{Average reward curves (max over steps) for an RPP-EMLP applied to the policy $\pi$ only, as well as an RPP-EMLP for both the policy $\pi$ and the critic $Q$. Mean and standard deviation taken over $4$ trials shown in the shaded region. Only minor performance gains are achieved if using RPP for the policy only, however this variant is more stable and can to train on Humanoid-v2 without diverging.}
\label{fig:model-free-results2}
\end{figure}
\section{Experimental Details}\label{app:experimental-details}

Here we present the training details of the models used in the paper. Experiments were run on private servers with NVIDIA Titan RTX and RTX $2080$ Ti GPUs. We estimate that all runs performed in the initial experimentation and final evaluation on the RL tasks used approximately $500$ GPU hours. 
The experiments on dynamical systems, CIFAR-$10$, and UCI data required an additional $200$ GPU hours.

\subsection{Synthetic Dataset Experiments (\ref{sec: dynamical-systems} and \ref{sec: post-equiv})}

The windy pendulum dataset is a variant of the double spring pendulum Hamiltonian system from \citet{finzi2021practical}. In addition to the Hamiltonian of the base system
\begin{equation*}
    H_0(x_1,x_2,p_1,p_2) = V(x_1,x_2)+T(p_1,p_2)
\end{equation*}
where $T(p_1,p_2) = \|p_1\|^2/2m_1 + \|p_2\|^2/2m_2$ and $V(x_1,x_2)=$
\begin{equation*}
    \tfrac{1}{2}k_1(\|x_1\|-\ell_1)^2 + \tfrac{1}{2}k_2(\|x_1-x_2\|-\ell_2)^2 + m_1g^\top x_1 + m_2g^\top x_2,
\end{equation*}
we add a perturbation $H_1(x_1,x_2,p_1,p_2) = -w^\top x_1-w^\top x_2$ that is the energy of the wind acting as a constant force pushing in the $w = [-8,-5,0]$ direction. Setting $H = H_0 +\epsilon H_1$, we can control the strength of the wind and we choose $\epsilon=0.01$. This perturbation breaks the $\mathrm{SO}(2)$ symmetry about the $z$ axis.

For the MLP, EMLP, and RPP we use $3$ layer deep $128$ hidden unit Hamiltonian neural networks \citep{greydanus2019hamiltonian} to fit the data using the rollouts of an ODE integrator \citep{chen2018neural} with an MSE loss on rollouts of length $5$ timesteps with $\Delta t=0.2$. For training we use $500$ trajectory chunks and use another $500$ for testing. We train all models in section \ref{sec: dynamical-systems} for $1000$ epochs, sufficient for convergence. The input and output representation for EMLP and RPP-EMLP is $V_{\mathrm{O}(3)}^4 \rightarrow \mathbb{R}$, where $V_{\mathrm{O}(3)}$ is the restricted representation from the standard representation of a 3D rotation matrix to the given group in question, like $\mathrm{SO}(2)$ for rotations about the $z$ axis. The input is $V_{\mathrm{O}(3)}^4$ because there are two point masses each of which has a $3$D vectors for position and for momentum. The scalar $\mathbb{R}$ output is the Hamiltonian function.

The Modified Inertia dataset is a small regression dataset off of the task also from \citet{finzi2021practical} for learning the moment of inertia matrix in 3D of a collection of $5$ point masses. For the base Inertia dataset, the targets are $\mathcal{I} = \sum_{i=1}^{5}m_i(x_i^\top x_iI-x_ix_i^\top)$ from the input tuples ${(m_i,x_i)}_{i=1}^5$. In order to break the equivariance of the dataset, we add an additional term so that the target is $y = \mathrm{vec}(\mathcal{I}+ 0.3\mathcal{I}^2\hat{z}\hat{z}^\top\mathcal{I})$ where $\hat{z}$ is the unit vector along the $z$ axis. The input and output representations for EMLP and RPP-EMLP on this problem are $(\mathbb{R}\oplus V)^5 \rightarrow V\otimes V$, representing the $5$ point masses and vectors mapping to matrices $V\otimes V$.

We use $1000$ train and test examples for the inertia datasets and we train for $500$ epochs. In both cases we use an Adam optimizer \citep{kingma2014adam} with a learning rate of $0.003$.

\subsection{Image and UCI experiments (\ref{sec: rpp-conv})}
We use the CIFAR-$10$ and UCI datasets, taken from \citet{krizhevsky2009learning} and \citet{Dua:2019} respectively. In Section \ref{sec: howandwhy} we train models on dynamical systems and CIFAR-$10$ and UCI regression data. 
For the CIFAR-$10$ experiments we use a convolutional neural network (and the equivalent MLP) with $9$ convolutional layers and $1$ fully connected layer, and max-pooling layers after the third and sixth convolutional layers. The channel sizes of the $9$ layers are, in order: $16, 16, 16, 32, 32, 32, 32, 32, 32$. 
We train for $200$ epochs using a cosine learning rate schedule with an initial learning rate of $0.05$ and the Adam optimizer. 

For the UCI tasks we use a small convolutional neural network, and the equivalent MLP, with $3$ convolutional layers and $1$ fully connected layer, with each convolutional layer having $32$ channels. Models are trained for $1000$ epochs using an Adam optimizer with a learning rate of $0.01$ and a cosine learning rate schedule.
 
\subsection{Model Free RL}
We train on the Mujoco locomotion tasks in the OpenAI gym environments \citep{brockman2016openai}. We follow the implementation details and hyperparameters from \citet{haarnoja2018soft1}, with a learned temperature function, stochastic policies, and double critics. Additionally we use the recommendation from \citet{andrychowicz2020matters} to initialize the last layer of the policy network with $100$x smaller weights, which we find slightly improves the performance of both RPP and the baseline. Additionally for RPP which can be less stable than standard SAC, we use the Adam betas $\beta_1=0.5$ and $\beta_2=0.999$ that are used in he GAN community \citep{miyato2018spectral} rather than the defaults. Training with the RPP $\pi$ and $Q$ functions on the Mujoco locomotion tasks takes about $8$ hours for $1$ million steps. 

We found it necessary to reduce the speed $\tau$ of the critic moving average to keep SAC stable on some of the environments, with values shown in \autoref{table:tau}. In general, higher $\tau$'s are favorable for learning quickly. Unfortunately we were not able to get SAC with an RPP Q function to train reliably on Humanoid, even after trying multiple values of $\tau$.

 \begin{table}[h]
\centering
    \begin{tabular}{ccccccc}
        & Walker2d & Hopper &  HalfCheetah& Swimmer &  Ant & Humanoid\\
         \midrule
         Baseline $\tau$& .005& .005& .005& .005&.005&.005\\
         RPP $\tau$& .004& .005& .005& .004& .005& \ding{55}\\
         \bottomrule
    \end{tabular}
    \vspace{2mm}
    \caption{Critic moving average speed $\tau$.}
    \label{table:tau}
    \vspace{-6mm}
\end{table}

\subsection{Transition Models for Mujoco}
We train the transition models on a dataset of $50000$ transitions which are composed of $5000$ trajectory chunks of length $10$. These trajectory chunks are sampled uniformly from the replay buffer collected over the course of training a standard SAC agent for $10^6$ steps on each of the environments. We train by minimizing the $\ell1$ norm of the rollout error over a $10$ step trajectory, and we evaluate on a holdout set of $50$ trajectories of length $100$.

The models are simple MLPs or RPPs mapping from the state and control actions to the state space, predicting the change in state,
\begin{equation*}
    x_{t+1} = x_t + \mathrm{NN}(x_{t},u_t).
\end{equation*}
For the MLPs and RPPs we use $2$ hidden layers of size $256$ as well as swish activations \citep{ramachandran2017searching}. We use a prior variance of $10^6$ in the equivariant subspace and $3$ in the non equivariant subspace. The RPP is a standard RPP-EMLP with the input representation $\rho_X \oplus \rho_U$ (concatenation of the representation of the state space and the action space), output representation $\rho_X$, and symmetry group described in \autoref{app:representations} the same as for the model free experiments. We train the transition models for $500$ epochs which takes about $45$ minutes for RPP compared to $15$ minutes for the standard MLPs.
 
 \section{Mujoco State and Action Representations}\label{app:representations}

Based on the state and action spaces of the Mujoco environments we describe in \autoref{app:spaces}, we define appropriate group representations on these spaces. Let $V$ be the base representation of the group acted upon by permutations for $\mathbb{Z}_n$ and by rotation matrices for $\mathrm{SO}(2)$, let $\mathbb{R}$ denote a scalar representation (of dimension $1$) that is unaffected by the transformations, and let $P$ be a pseudoscalar representation (of dimension $1$) that transforms by the sign of the permutation. For $\mathbb{Z}_2$, $P$ takes the values $1$ and $-1$ and acts by negating the values when a flip or L/R reflection is applied.

\begin{table}[h]
\caption{Mujoco Locomotion State and Action Representations used for RPP-EMLP}
\begin{center}
\begin{tabular}{c ccc}
\midrule
Env &State Representation & Action Rep & Group\\
\midrule
Hopper &$\mathbb{R}\oplus P^5\oplus \mathbb{R}\oplus P^4$& $P^3$ & $\mathbb{Z}_2$\\
Swimmer &$\mathbb{R}\oplus P_\leftrightarrow \oplus (P_\leftrightarrow \otimes V_\updownarrow) \oplus (\mathbb{R}\oplus P)^2\oplus (P_\leftrightarrow \otimes V_\updownarrow)$& $P_\leftrightarrow \otimes V_\updownarrow$ & $\mathbb{Z}_2^\leftrightarrow \times \mathbb{Z}_2^\updownarrow$\\
HalfCheetah & $\mathbb{R}\oplus P^8 \oplus \mathbb{R} \oplus P^7$&$P^6$ & $\mathbb{Z}_2$\\
Walker2d &$\mathbb{R}^2\oplus V^3\oplus\mathbb{R}^3\oplus V^3$ & $V^3$& $\mathbb{Z}_2$\\
Ant & $\mathbb{R}^5\oplus V^2\oplus \mathbb{R}^6\oplus V^2$& $V^2$& $\mathbb{Z}_4$\\
Humanoid &$\mathbb{R}\oplus V_{\mathrm{SO(3)}}^{\otimes2}\oplus \mathbb{R}^{17} \oplus V_{\mathrm{SO(3)}}^2 \oplus \mathbb{R}^{17}$ & $\mathbb{R}^{17}$& $\mathrm{SO}(2)$\\
\end{tabular}
\end{center}
\label{tab:representations}
\end{table}

From the raw state and action spaces listed in \autoref{app:spaces}, we convert quaternions to 3D rotation matrices for Humanoid and Ant, and we reorder elements to group together left/right pairs for Walker2d and Swimmer. The representations of these transformed state and action vectors are shown in \autoref{tab:representations}. Note that $V^3$ denotes $V\oplus V\oplus V = V^{\oplus 3}$, and is simply the concatenation of $3$ copies of $V$ as $\mathbb{R}^3$ would be $3$ copies of $\mathbb{R}$. This is not to be confused with powers of the tensor product, $V^{\otimes 3} = V\otimes V\otimes V$. For Humanoid, we denote the restricted representation of $3$D rotation matrices restricted to the $\mathrm{SO}(2)$ rotations about the $z$ axis as $V_{\mathrm{SO}(3)}$.

\section{Mujoco State and Action Spaces}\label{app:spaces}
In order to build symmetries into the state and action representations for Mujoco environments, we need to have a detailed understanding of what the state and action spaces for these environments represent. As these spaces are not well documented, for each of the Mujoco environments we experimented in the simulator and identified the meanings of the state vectors in Tables \ref{tab:ant}, \ref{tab:humanoid-states}, \ref{tab:humanoid-action}, \ref{tab:swimmer}, \ref{tab:walker}, \ref{tab:hopper}, and \ref{tab:halfcheetah}. We hope that these detailed descriptions can be useful to other researchers.

\begin{table}[!htb]
    \begin{minipage}{.5\linewidth}
\caption{Hopper-v2 State and Action Spaces}
\begin{center}
\begin{tabular}{c|c}
    \hline
    \hline
     \multirow{12}{*}{State Space} & X (Unobserved)\\
    \cline{2-2}
    & Y\\
    \cline{2-2}
    & Orientation Angle \\
    \cline{2-2}
    & Hip Angle \\
    \cline{2-2}
    & Knee Angle \\
    \cline{2-2}
    & Ankle Angle \\
    \cline{2-2}
    & X Velocity \\
    \cline{2-2}
    & Y Velocity \\
    \cline{2-2}
    & Orientation Angular Velocity \\
    \cline{2-2}
    & Hip Angular Velocity\\
    \cline{2-2}
    & Knee Angular Velocity\\
    \cline{2-2}
    & Ankle Angular Velocity\\
    \hline
    \multirow{3}{*}{Action Space}& Hip \\
    \cline{2-2}
    & Knee \\
    \cline{2-2}
    & Ankle \\
    \hline\hline
\end{tabular}
\end{center}
\label{tab:hopper}
    \end{minipage}%
    \begin{minipage}{.5\linewidth}
      \caption{Swimmer-v2 State and Action Spaces}
\begin{center}
\begin{tabular}{c|c}
    \hline
    \hline
     \multirow{10}{*}{State Space} & X (Unobserved)\\
    \cline{2-2}
    & Y (Unobserved)\\
    \cline{2-2}
    & Orientation Angle \\
    \cline{2-2}
    & Head Joint Angle \\
    \cline{2-2}
    & Tail Joint Angle \\
    \cline{2-2}
    & X Velocity \\
    \cline{2-2}
    & Y Velocity \\
    \cline{2-2}
    & Orientation Angular Velocity \\
    \cline{2-2}
    & Head Joint Angular Velocity \\
    \cline{2-2}
    & Tail Joint Angular Velocity \\
    \hline
    \multirow{2}{*}{Action Space}& Head Joint \\
    \cline{2-2}
    & Tail Joint \\
    \hline
    \hline
\end{tabular}
\end{center}
\label{tab:swimmer}
    \end{minipage} 
\end{table}

\begin{table}[!htb]
    \begin{minipage}{.5\linewidth}
    \caption{HalfCheetah-v2 State and Action Spaces}
\begin{center}
\begin{tabular}{c|c}
    \hline
    \hline
     \multirow{18}{*}{State Space} & X (Unobserved)\\
    \cline{2-2}
    & Y \\
    \cline{2-2}
    & Orientation Angle \\
    \cline{2-2}
    & Rear Hip Angle \\
    \cline{2-2}
    & Rear Knee Angle \\
    \cline{2-2}
    & Rear Ankle Angle \\
    \cline{2-2}
    & Front Hip Angle \\
    \cline{2-2}
    & Front Knee Angle \\
    \cline{2-2}
    & Front Ankle Angle \\
    \cline{2-2}
    & X Velocity \\
    \cline{2-2}
    & Y Velocity \\
    \cline{2-2}
    & Orientation Angular Velocity \\
    \cline{2-2}
    & Rear Hip Angular Velocity\\
    \cline{2-2}
    & Rear Knee Angular Velocity\\
    \cline{2-2}
    & Rear Ankle Angular Velocity\\
    \cline{2-2}
    & Front Hip Angular Velocity\\
    \cline{2-2}
    & Front Knee Angular Velocity\\
    \cline{2-2}
    & Front Ankle Angular Velocity\\
    \hline
    \multirow{6}{*}{Action Space} & Rear Hip \\
    \cline{2-2}
    & Rear Knee \\
    \cline{2-2}
    & Rear Ankle \\
    \cline{2-2}
    & Front Hip \\
    \cline{2-2}
    & Front Knee \\
    \cline{2-2}
    & Front Ankle \\
    \hline\hline
\end{tabular}
\end{center}
\label{tab:halfcheetah}
    \end{minipage} 
    \begin{minipage}{.5\linewidth}
    \caption{Walker2d-v2 State and Action Spaces}
\begin{center}
\begin{tabular}{c|c}
    \hline
    \hline
     \multirow{18}{*}{State Space} & X (Unobserved)\\
    \cline{2-2}
    & Y \\
    \cline{2-2}
    & Orientation Angle \\
    \cline{2-2}
    & Right Hip Angle \\
    \cline{2-2}
    & Right Knee Angle \\
    \cline{2-2}
    & Right Ankle Angle \\
    \cline{2-2}
    & Left Hip Angle \\
    \cline{2-2}
    & Left Knee Angle \\
    \cline{2-2}
    & Left Ankle Angle \\
    \cline{2-2}
    & X Velocity \\
    \cline{2-2}
    & Y Velocity \\
    \cline{2-2}
    & Orientation Angular Velocity \\
    \cline{2-2}
    & Right Hip Angular Velocity\\
    \cline{2-2}
    & Right Knee Angular Velocity\\
    \cline{2-2}
    & Right Ankle Angular Velocity\\
    \cline{2-2}
    & Left Hip Angular Velocity\\
    \cline{2-2}
    & Left Knee Angular Velocity\\
    \cline{2-2}
    & Left Ankle Angular Velocity\\
    \hline
    \multirow{6}{*}{Action Space}& Right Hip \\
    \cline{2-2}
    & Right Knee \\
    \cline{2-2}
    & Right Ankle \\
    \cline{2-2}
    & Left Hip \\
    \cline{2-2}
    & Left Knee \\
    \cline{2-2}
    & Left Ankle \\
    \hline\hline
\end{tabular}
\end{center}
\label{tab:walker}
    \end{minipage} 
    
\end{table}

\begin{table}[!htb]
    \begin{minipage}{.5\linewidth}
    \caption{Ant-v2 State and Action Spaces}
\begin{center}
\begin{tabular}{c|c}
    \hline
    \hline
     \multirow{13}{*}{State Space} & X (Unobserved)\\
    \cline{2-2}
    & Y (Unobserved)\\
    \cline{2-2}
    & Z\\
    \cline{2-2}
    & Orientation Quaternion ($4D$) \\
    \cline{2-2}
    & Limb $2$ Left/Right \\
    \cline{2-2}
    & Limb $2$ Up/Down \\
    \cline{2-2}
    & Limb $3$ Left/Right \\
    \cline{2-2}
    & Limb $3$ Up/Down \\
    \cline{2-2}
    & Limb $4$ Left/Right \\
    \cline{2-2}
    & Limb $4$ Up/Down \\
    \cline{2-2}
    & Limb $1$ Left/Right \\
    \cline{2-2}
    & Limb $1$ Up/Down \\
    \cline{1-2}
    \multirow{8}{*}{Action Space}
    & Limb $1$ Left/Right \\
    \cline{2-2}
    & Limb $1$ Up/Down \\
    \cline{2-2}
    & Limb $2$ Left/Right \\
    \cline{2-2}
    & Limb $2$ Up/Down \\
    \cline{2-2}
    & Limb $3$ Left/Right \\
    \cline{2-2}
    & Limb $3$ Up/Down \\
    \cline{2-2}
    & Limb $4$ Left/Right \\
    \cline{2-2}
    & Limb $4$ Up/Down \\
    \hline
\end{tabular}
\end{center}
\label{tab:ant}

\caption{Humanoid-v2 Action Space}
\begin{center}
\begin{tabular}{c|c}
    \hline \hline
    \multirow{17}{*}{Action Space} & Torso Forward/Backward \\
    \cline{2-2}
    & Torso Z \\
    \cline{2-2}
    & Torso Left/Right \\
    \cline{2-2}
    & Right Hip Left/Right \\
    \cline{2-2}
    & Right Hip Up/Down \\
    \cline{2-2}
    & Right Hip Front/Back \\
    \cline{2-2}
    & Right Knee Front/Back \\
    \cline{2-2}
    & Left Hip Left/Right \\
    \cline{2-2}
    & Left Hip Up/Down \\
    \cline{2-2}
    & Left Hip Front/Back \\
    \cline{2-2}
    & Left Knee Front/Back \\
    \cline{2-2}
    & Right Shoulder Left/Right \\
    \cline{2-2}
    & Right Shoulder Front/Back \\
    \cline{2-2}
    & Right Elbow Front/Back \\
    \cline{2-2}
    & Left Shoulder Left/Right \\
    \cline{2-2}
    & Left Shoulder Front/Back \\
    \cline{2-2}
    & Left Elbow Front/Back \\
    \hline \hline
\end{tabular}
\end{center}
\label{tab:humanoid-action}
    \end{minipage} 
    \begin{minipage}{.4\linewidth}
\caption{Humanoid-v2 State Space}
\begin{center}
\begin{tabular}{c|c}
    \hline \hline
    \multirow{21}{*}{State Space (Position)} & X (Unobserved)\\
    \cline{2-2}
    & Y (Unobserved)\\
    \cline{2-2}
    & Z\\
    \cline{2-2}
    & Orientation Quaternion ($4D$) \\
    \cline{2-2}
    & Torso Z \\
    \cline{2-2}
    & Torso Forward/Backward \\
    \cline{2-2}
    & Torso Left/Right \\
    \cline{2-2}
    & Right Hip Left/Right \\
    \cline{2-2}
    & Right Knee Left/Right \\
    \cline{2-2}
    & Right Hip Up/Down \\
    \cline{2-2}
    & Right Knee Up/Down \\
    \cline{2-2}
    & Left Hip Left/Right \\
    \cline{2-2}
    & Left Knee Left/Right \\
    \cline{2-2}
    & Left Hip Up/Down \\
    \cline{2-2}
    & Left Knee Up/Down \\
    \cline{2-2}
    & Right Shoulder Left/Right \\
    \cline{2-2}
    & Right Shoulder Up/Down \\
    \cline{2-2}
    & Right Elbow Left/Right \\
    \cline{2-2}
    & Left Shoulder Left/Right \\
    \cline{2-2}
    & Left Shoulder Up/Down \\
    \cline{2-2}
    & Left Elbow Left/Right \\
    \cline{1-2}
    
    \multirow{22}{*}{State Space (Velocity)}
    & Body Linear Velocity ($3D$)\\
    \cline{2-2}
    & Body Angular Velocity ($3D$) \\
    \cline{2-2}
    & Torso Z \\
    \cline{2-2}
    & Torso Forward/Backward \\
    \cline{2-2}
    & Torso Left/Right \\
    \cline{2-2}
    & Right Hip Left/Right \\
    \cline{2-2}
    & Right Knee Left/Right \\
    \cline{2-2}
    & Right Hip Up/Down \\
    \cline{2-2}
    & Right Knee Up/Down \\
    \cline{2-2}
    & Left Hip Left/Right \\
    \cline{2-2}
    & Left Knee Left/Right \\
    \cline{2-2}
    & Left Hip Up/Down \\
    \cline{2-2}
    & Left Knee Up/Down \\
    \cline{2-2}
    & Right Shoulder Left/Right \\
    \cline{2-2}
    & Right Shoulder Up/Down \\
    \cline{2-2}
    & Right Elbow Left/Right \\
    \cline{2-2}
    & Left Shoulder Left/Right \\
    \cline{2-2}
    & Left Shoulder Up/Down \\
    \cline{2-2}
    & Left Elbow Left/Right \\
    \hline\hline
\end{tabular}
\label{tab:humanoid-states}
\end{center}
    \end{minipage} 
\end{table}

\end{document}